\pdfoutput=1
\documentclass[11pt]{article}
\usepackage{lipsum}    
\usepackage{amsmath}

\usepackage{amsmath}
\usepackage[final]{acl}
\usepackage{amssymb}
\usepackage{mathrsfs}
\usepackage{calligra}
\usepackage{graphicx}
\usepackage[table]{xcolor}
\usepackage{multirow}
\usepackage{xcolor}
\definecolor{darkgreen}{rgb}{0.0, 0.5, 0.0}

\usepackage{pifont}
\usepackage{times}
\usepackage{latexsym}
\usepackage{booktabs} 
\usepackage{listings}
\usepackage{tikz}
\usepackage[T1]{fontenc}
\usepackage[utf8]{inputenc}
\usepackage{booktabs}
\usepackage{array}

\usepackage{times}
\usepackage{latexsym}
\usepackage{multicol}
\usepackage{subcaption}

\usepackage[table]{xcolor}
\usepackage{colortbl}
\usepackage{geometry}      % Optional, for better page layout
\usepackage{adjustbox}
\usepackage{multirow}
\usepackage{microtype}

\usepackage{inconsolata}
\usepackage{booktabs}
\usepackage{array}
\usepackage{multirow}
\usepackage{hyperref}

\usepackage{microtype}

\usepackage{inconsolata}

\usepackage{graphicx}
\usepackage{bbm}
\usepackage[breakable]{tcolorbox}

\usepackage{algorithm}
\usepackage{algpseudocode}
\usepackage{float}
\usepackage[colorinlistoftodos,prependcaption,textsize=tiny]{todonotes}

\makeatletter
\renewcommand{\@makefnmark}{\hbox{\@textsuperscript{\normalfont\color{black}\@thefnmark}}}
\makeatother

\title{Leveraging the Cross-Domain \& Cross-Linguistic Corpus for Low Resource NMT: A Case Study On Bhili-Hindi-English Parallel Corpus}

\author{
  Pooja Singh\textsuperscript{1} \quad
  Shashwat Bhardwaj\textsuperscript{3} \quad
  Vaibhav Sharma\textsuperscript{1} \quad
  Sandeep Kumar\textsuperscript{1,2,3} \\
  \textsuperscript{1}Department of Electrical Engineering, Indian Institute of Technology Delhi \\
  \textsuperscript{2}Bharti School of Telecommunication, Indian Institute of Technology Delhi \\
  \textsuperscript{3}Yardi School of Artificial Intelligence, Indian Institute of Technology Delhi \\
  New Delhi, India \\
  \texttt{\{eez228470,\,aiy237528,\,eey257541,\,ksandeep\}@iitd.ac.in}
}

\begin{document}
\maketitle

\begin{abstract}
The linguistic diversity of India  poses significant machine translation challenges, especially for underrepresented tribal languages like Bhili, which lack high-quality linguistic resources. This paper addresses the gap by introducing Bhili-Hindi-English Parallel Corpus (BHEPC), the first and largest parallel corpus worldwide comprising 110,000 meticulously curated sentences across Bhili, Hindi, and English. The corpus was created with the assistance of expert human translators. BHEPC  spans critical domains such as education, administration, and news,  establishing a valuable benchmark for research in low resource machine translation. To establish a comprehensive Bhili Machine Translation benchmark, we evaluated  a wide range of proprietary and open-source Multilingual Large Language Models (MLLMs) on bidirectional translation tasks between English/Hindi and Bhili. Comprehensive evaluation demonstrates that the fine-tuned NLLB-200 distilled 600M variant model outperforms others, highlighting the potential of multilingual models in low resource scenarios. Furthermore, we investigated the generative translation capabilities of multilingual LLMs on BHEPC using in-context learning, assessing performance under cross-domain generalization and quantifying distributional divergence. 
 This work bridges a critical resource gap and promotes inclusive natural language processing technologies for low-resource and marginalized languages globally.
\end{abstract}

\section{Introduction}

India is a linguistically diverse country, with over $1,369$ distinct mother tongues reported in the $2011$ census and $22$ constitutionally recognized languages under the $8^{th}$ Schedule.
However, despite this linguistic richness, many indigenous and tribal languages remain critically endangered due to an acute scarcity of digitized linguistic resources and parallel corpora. Indigenous languages are deeply intertwined with cultural heritage, robust translation systems are imperative not only for effective communication and social inclusion but also for equitable access to government services~\cite{nekoto-etal-2020-participatory}. Effective translation supports crucial national activities ranging from the dissemination of policy directives and welfare schemes to judicial processes and educational initiatives, thereby reinforcing national cohesion within India's intricate multilingual landscape~\cite{haddow2022survey}.

Efforts to improve machine translation (MT) for Indic languages have accelerated, with early initiatives like the Hindi-English MT challenge at WMT’14~\cite{bojar2014findings} facilitating subsequent benchmarks for Gujarati-English~\cite{barrault2019findings} and Tamil-English~\cite{akhbardeh2021findings}. Recent large-scale efforts such as Workshop on Asian Translation(WAT) ~\cite{nakazawa2021proceedings}, FLORES 101~\cite{goyal2022flores} NLLB~\cite{costa2022no} and INDICGENBENCH~\cite{singh2024indicgenbench} 
have expanded linguistic coverage significantly by incorporating multiple Indic languages into MT benchmarks. 
Parallel advances  in Large Language Models (LLMs)~\cite{achiam2023gpt,tay2022ul2,team2023gemini} have revolutionized text generation and translation tasks. Howerver, Contemporary multilingual models, predominantly trained on high-resource languages, inherently carry cultural biases, leading to suboptimal performance on underrepresented languages. 
In this context, the Bhili language, spoken by approximately 13 million people,\footnote{\href{https://en.wikipedia.org/wiki/Bhili_language}{https://en.wikipedia.org/wiki/Bhili\_language}} and written in the Devanagari script, is notably underserved. The lack of parallel corpora has severely impeded the development of effective MT systems for Bhili.
 
 To address this gap, we present the Bhili-Hindi-English Parallel Corpus (BHEPC), a meticulously curated, community-driven, high-quality  corpus comprising 110,000 aligned sentences in Bhili, Hindi, and English. Leveraging richer linguistic resources available in Hindi and English, BHEPC enables the exploration of transfer learning and cross-lingual methodologies in an extremely low-resource setting.
Although this study focuses on Bhili, the proposed corpus construction workflow, starting from a seed set and iteratively expanded through model-assisted generation with native speaker post-editing, offers a scalable methodology that can be adapted to other endangered languages with minimal digital presence. Building on this foundation, our primary contributions are:
 
% Moreover, this study establishes a comprehensive evaluation benchmark to assess multilingual translation capabilities of prominent  open-source and proprietary LLMs specifically targeting Bhili translation tasks, our evaluation provides critical insights into model performance, promoting linguistic inclusivity and digital empowerment for Bhili speakers. 

% %\vspace{-0.75mm}
\begin{itemize}
   
    \item We introduce the Bhili-Hindi-English Parallel Corpus (BHEPC), the first large-scale, gold-standard parallel corpus for Bhili, comprising 110,000 sentences meticulously curated by native speakers through community-driven efforts.
% %\vspace{-0.75mm}

\item We extensively benchmark the multilingual translation capabilities of state-of-the-art open-source \& proprietary models such as mT5, Qwen3, DeepSeek-V3, Gemma-2-9B, Mistral-7B-v0.1, BLOOMZ, Llama-2, Llama-3, Llama-4-Scout-17B-16E, Gemini 2.0 Flash, Gemini 2.5 Flash, GPT-3.5 Turbo, GPT-4o-0513, and GPT-4.5 across various model sizes across four translation directions: Hindi$\leftrightarrow$Bhili and English$\leftrightarrow$Bhili.
% %\vspace{-0.75mm}
\item  We analyze cross-domain generalization and quantify distributional divergence across translation directions using Jensen-Shannon Divergence (JSD), while also proposing a hybrid seed-and-post-editing workflow that reduces manual translation effort and provides a scalable template for other endangered languages.  
\end{itemize}

\section{Background}

Large Language models heavily rely on large annotated corpora, unintentionally favoring high-resource languages, leaving low-resource languages, particularly several regional languages of India, notably underrepresented in computational models and digital repositories. 
This discrepancy necessitates 
targeted initiatives for creating community-driven, gold-standard datasets to capture cultural nuances and ensure digital equity and cultural preservation.

The introduction of Transformer-based models~\cite{vaswani2017attention}, has brought about a paradigm change in neural machine translation (NMT). 
Despite these advances, translating extremely low-resource regional Indic languages such as Bhili, remains challenging due to inadequate parallel and monolingual data. 
While transfer learning~\cite{zoph2016transfer,chen2023improving}  and multilingual NMT ~\cite{johnson2017google,dabre2020survey} partly mitigate data scarcity, their effectiveness is still limited by reliance on parallel corpora.  

Concurrent research has also focused on building natural language understanding models~\cite{kakwani2020indicnlpsuite,khanuja2021muril}  and comprehensive evaluation datasets~\cite{doddapaneni2023towards,mhaske2023naamapadam} for Indic languages, thus facilitating systematic benchmarking and comparison.
Recent multilingual models like IndicTrans2~\cite{gala2023aswanth} and MuRIL~\cite{khanuja2021muril} offer improved translation and natural language understanding capabilities, leveraging synthetic data and transliteration to compensate for resource constraints. 
Meanwhile, large language models~\cite{costa2022no, arivazhagan2019missing,maurya2024charspan} demonstrate significant potential for cross-lingual transfer, yet their applicability to translation tasks without direct parallel supervision remains underexplored. Collectively, these advancements highlight the necessity of developing robust community-driven gold standard datasets, evaluation benchmarks, and specialized architectures to enhance translation for severely underrepresented languages like Bhili.

% trained on expansive multilingual corpora exhibit considerable potential for cross-lingual transfer, yet their efficacy in translation settings without direct parallel supervision remains underexplored. Altogether, these contributions underscore the importance of expanding datasets, establishing robust evaluation benchmarks, and devising specialized architectures to advance low-resource translation efforts, especially for underrepresented languages like Bhili.
% culturally grounded methods, community-driven data creation

\section{Dataset}

% Despite growing interest in low-resource machine translation (MT) from both researchers~\cite{haddow2022survey} and native speaker communities~\cite{nekoto-etal-2020-participatory}, extremely low-resource languages remain underrepresented in existing MT benchmarks and systems. 

The acute scarcity of publicly available parallel corpora continues to hinder the development of neural MT models for languages such as Bhili. Given Bhili’s linguistic and cultural significance, we address this gap by curating a large-scale, gold-standard parallel corpus translated by native speakers through community-driven efforts, constructing robust evaluation benchmarks, and leveraging multilingual models to exploit linguistic similarities across Indic languages.

\subsection{Corpus Details}
We introduce the Bhili-Hindi-English Parallel Corpus (BHEPC), the first large-scale, human-curated, gold-standard dataset developed through community efforts to support Bhili language translation.
The corpus comprises 1,08,000 training, 1000 validation, and 1,000 test sentences, across various domains, including education, administration, and mass media. Hindi source sentences were derived from established resources and meticulously translated into Bhili by language experts, thereby ensuring precise and contextually appropriate renderings. Hindi source sentences were primarily drawn from the Bharat Parallel Corpus Collection (BPCC)~\cite{gala2023aswanth}, supplemented by publicly available government documents from Legislative Assembly Speeches~\cite{siripragada2020multilingual}, PMIndia corpus~\cite{haddow2020pmindia}, and NCERT Textbooks\footnote{\href{https://ncert.nic.in/textbook.php}{https://ncert.nic.in/textbook.php}}.  Data collection and manual translation efforts, spanning May 2024 to March 2025, involved an average 10 professional translators contributing a total of 27,000 hours. The dataset includes Bhili-Hindi-English tripartite parallel sentences, systematically screened to remove personally identifiable information, hate speech, and redundancy before segmentation into sentence pairs, more details are provided in Appendix~\ref{app:preprocessing}. 

In addition to structural details, BHEPC was deliberately curated to encode cultural and social dimensions; Appendix~\ref{app:culture} outlines how orthography, idioms, and community practices were preserved to make the dataset both a linguistic resource and a cultural benchmark.
Table~\ref{table:dataset_statistics} presents detailed corpus statistics, including sentence counts, token distributions, and vocabulary diversity across language pairs.

% \begin{table}[!htbp]
%     \centering
%     \caption{Statistics of Datasets}
%     \label{table:dataset_statistics}
%     \begin{tabular}{l c c c c}
%         \toprule
%         & Lang. & Train & Test & Dev \\
%         \midrule
%         \#Sent. & & 1,08,000 & 1,000 & 1,000 \\
%          \midrule
%         \#Tokens & eng & 22,21,303 & 20,413 & 20,976 \\
%         & hin & 23,57,588 & 24,062 & 24,546 \\
%         & bhb & 23,83,485 & 26.017 & 26,326\\
%          \midrule
%         \#Types & eng & 1,08,012&  6,478 & 6,589 \\
%         & hin &  1,06,749 & 5,324 & 5,834 \\
%         & bhb & 2,00,248 & 7,648 & 7,851 \\
%         \bottomrule
%     \end{tabular}
% \end{table}

\begin{table}[!t]
    \centering
    \begin{tabular}{l c c c c}
        \toprule
        & Lang. & Train & Test & Dev \\
        \midrule
        \#Sent. & & 1,08,000 & 1,000 & 1,000 \\
         \midrule
        \#Tokens & eng & 22,21,303 & 20,413 & 20,976 \\
        & hin & 23,57,588 & 24,062 & 24,546 \\
        & bhb & 23,83,485 & 26.017 & 26,326\\
         \midrule
        \#Types & eng & 1,08,012&  6,478 & 6,589 \\
        & hin &  1,06,749 & 5,324 & 5,834 \\
        & bhb & 2,00,248 & 7,648 & 7,851 \\
        \bottomrule
    \end{tabular}
    \caption{Statistics of Datasets}
    \label{table:dataset_statistics}
\end{table}

% \begin{table}[!t]
%     \centering
%     % tighten vertical spacing
%     \setlength{\aboverulesep}{0pt}
%     \setlength{\belowrulesep}{0pt}
%     \renewcommand{\arraystretch}{0.93} % < 1.0 tightens row height

%     \begin{tabular}{@{}lcccc@{}} % @{} trims left/right padding
%         \toprule
%         & Lang. & Train & Test & Dev \\
%         \midrule
%         \#Sent. & & 1,08,000 & 1,000 & 1,000 \\
%         \midrule
%         \#Tokens & eng & 22,21,303 & 20,413 & 20,976 \\
%         & hin & 23,57,588 & 24,062 & 24,546 \\
%         & bhb & 23,83,485 & 26.017 & 26,326\\
%         \midrule
%         \#Types & eng & 1,08,012&  6,478 & 6,589 \\
%         & hin &  1,06,749 & 5,324 & 5,834 \\
%         & bhb & 2,00,248 & 7,648 & 7,851 \\
%         \bottomrule
%     \end{tabular}
%     \caption{Statistics of Datasets}
%     \label{table:dataset_statistics}
% \end{table}

\begin{table*}[!t]
\centering
\resizebox{0.98\textwidth}{!}{%
\begin{tabular}{lcccccccccccc}
\toprule
\textbf{Model (LLM)} & \multicolumn{3}{c}{\textbf{hin-bhb}} & \multicolumn{3}{c}{\textbf{bhb-hin}} & \multicolumn{3}{c}{\textbf{eng-bhb}} & \multicolumn{3}{c}{\textbf{bhb-eng}} \\
\cmidrule(lr){2-4} \cmidrule(lr){5-7} \cmidrule(lr){8-10} \cmidrule(lr){11-13}
\textbf{In-Context Ex.} & 0 & 5 & 10 & 0 & 5 & 10 & 0 & 5 & 10 & 0 & 5 & 10 \\
\midrule
Llama -2-7B & 5.61 & 12.87 & 14.89 & 20.07 & 19.57 & 15.06 & 0.43 & 12.92 & 11.67 & 11.45 & 9.25 & 15.15 \\
Llama -3.2-1B  & 4.78 & 9.50 & 11.64 & 19.11 & 13.50 & 12.14 & 0.36 & 7.08 & 11.26 & 10.76 & 14.93 & 12.82 \\
Llama -3-8B  & 7.50 & 13.50 & 15.65 & 21.23 & 20.50 & 16.80 & 0.55 & 13.50 & 13.80 & 12.00 & 16.50 & 17.50 \\
Llama-4-Scout-17B-16E & \colorbox{cyan!30}{13.89} & 15.76  & \colorbox{gray!30}{17.41}  & 22.45  & \colorbox{gray!30}{25.64} & 17.32 & \colorbox{cyan!30}{6.35} & \colorbox{cyan!30}{16.78} & \colorbox{cyan!30}{23.64} & \colorbox{gray!30}{24.68} & \colorbox{cyan!30}{25.67} & 22.29 \\
BLOOMZ-560M & 5.67 & \colorbox{gray!30}{17.67} & \colorbox{cyan!25}{18.76} & 23.07 & 23.89 & 25.37 & 0.37 & 9.88 & 7.50 & 13.70 & 18.35 & 13.87 \\
BLOOMZ-3.1B & 6.35 & 12.19 & 14.00 & \colorbox{cyan!30}{23.87} & 25.42 & \colorbox{cyan!30}{27.04} & 0.64 & 10.13 & 12.45 & 19.01 & 18.65 & 20.12 \\
BLOOMZ-7B1 & \colorbox{gray!30}{12.58} & 13.54 & 15.24 & \colorbox{cyan!30}{25.21} & \colorbox{cyan!30}{26.33} & \colorbox{gray!30}{26.15} & 0.81 & 11.12 & \colorbox{gray!30}{15.54} & 21.34 & 22.98 & \colorbox{gray!30}{23.87} \\
Mistral-7B-v0.1 & 4.61 & 14.86 & 12.87 & 13.89 & 12.50 & 17.23 & 0.76 & 10.87 & 11.43 & 12.89 &  13.25& 14.21 \\
Gemma-2-9B & 9.89 & \colorbox{cyan!30}{19.03} & 16.34 & 17.51 & 19.64 & 15.03 & \colorbox{gray!30}{5.27} & \colorbox{gray!30}{13.53} & 11.90 & \colorbox{cyan!30}{30.50} & \colorbox{gray!30}{23.25} & \colorbox{cyan!30}{24.02} \\

\rowcolor{pink!30}
Gemini 2.0 Flash & 21.17 & 23.44 & 24.56 & 33.01 & 34.78 & 36.01 & 16.82 & 18.60 & 19.13 & 35.62 & 37.56 & 38.37 \\

\rowcolor{pink!40}
Gemini 2.5 Flash & 24.19 & 25.05 & 26.74 & 34.09 & 35.17 & 35.89 & 19.35 & 21.30 & 23.54 & 37.56 & 38.89 & 39.12 \\

\rowcolor{pink!50}
GPT-3.5 Turbo & 21.53 & 26.31 & 28.32 & 39.12 & 40.32 & 42.13 & 20.39 & 21.78 & 23.01 & 38.03 & 40.15 & 41.91 \\

\rowcolor{pink!70}
GPT-4o-0513 & 24.01 & 28.33 & 28.89 & 43.09 & 44.19 & 45.32 & 22.19 & 22.76 & 24.89 & 40.51 & 41.36 & 43.57 \\

\rowcolor{pink!90}
GPT-4.5 & 27.47 & 29.75 & 31.23 & 45.72 & 46.68 & 49.65 & 23.35 & 25.67 & 27.12 & 42.16 & 43.78 & 45.16 \\

\bottomrule
\end{tabular}%
}
\caption{chrF++ scores with varying in-context examples across four different translation directions. For open-source models, the best scores are in green and second-best in grey. Proprietary models (GPT-4.5 \& Gemini) dominate overall, while smaller open source models remain competitive in low-resource settings.}
\label{tab:shot}
\end{table*}
\subsection{Data Preprocessing}
We applied extensive preprocessing to improve data quality and linguistic consistency. This included normalizing Bhili homophones, removing extraneous characters, converting English text to lowercase, and enforcing strict de-duplication to  avoid the repetition of nearly identical sentences, keeping only one example from sets of very similar sentences.
We also set limits on sentence length, rejecting those with fewer than 6 or more than 80 words. Furthermore, to eliminate redundant sentence pairs, we applied cosine similarity-based filtering to those with nearly identical source and target segments. Tokenization was performed using the SentencePiece model, ensuring uniform segmentation across language pairs. The English dataset was generated by translating Hindi sentences from the specified resources using the IndicTrans2 model. Further details on preprocessing steps, quality control, and validation procedures are provided in Appendix~\ref{app:preprocessing}.

%\vspace{-2mm}
\subsection{Evaluation Dataset \& Metrics }

To assess multilingual language models for the Bhili language effectively across all the translation directions, we curated a high-quality evaluation dataset. A stratified sampling approach was employed to extract a representative set of
sentences, ensuring proportional coverage across
different domains while avoiding redundancy.
Sentences from various source domains were combined to construct a balanced evaluation dataset. The remaining corpus was partitioned into training and validation sets in a 99:1 ratio while preserving domain distribution. 
We also curated a domain-specific evaluation set to establish robust benchmarks for evaluating cross-domain adaptability. 
This included 288 sentences from the NCERT domain, 487 from the Govt/PMI domain, and 1,063 from the mass media domain. Expert translators provided gold-standard translations to ensure high-quality reference data. This benchmark dataset serves as a foundation for evaluating fine-tuning strategies, cross-lingual transfer learning, and domain generalization on Bhili MT. For evaluating translation performance in low-resource language (LRL) settings, we report the chrF++ and a sentence-level variant of BLEU, spBLEU which is more robust than corpus-level metrics in scenarios with limited reference translations following prior work~\cite{khiu2024predicting}. We complement these automatic evaluations with detailed human judgments and inter-annotator agreement analysis, presented in Section~\ref{sec:human_eval}.

% \begin{figure*}[!htbp]
%     \centering
%     % Make sure the image is large enough to cover both columns
%     \includegraphics[width=0.9\textwidth, height=3.5cm]{Images/combined_translation_plots.pdf}
%     \caption{chrF++ performance trends of LLMs across in-context examples (0, 5, 10) shots for four translation directions: Hindi$\leftrightarrow$Bhili and English$\leftrightarrow$Bhili direction. }
%     \label{fig:curves models}
% \end{figure*}

\begin{figure*}[!t]
    \centering
    % Make sure the image is large enough to cover both columns
    \includegraphics[width=\textwidth, height=4 cm]{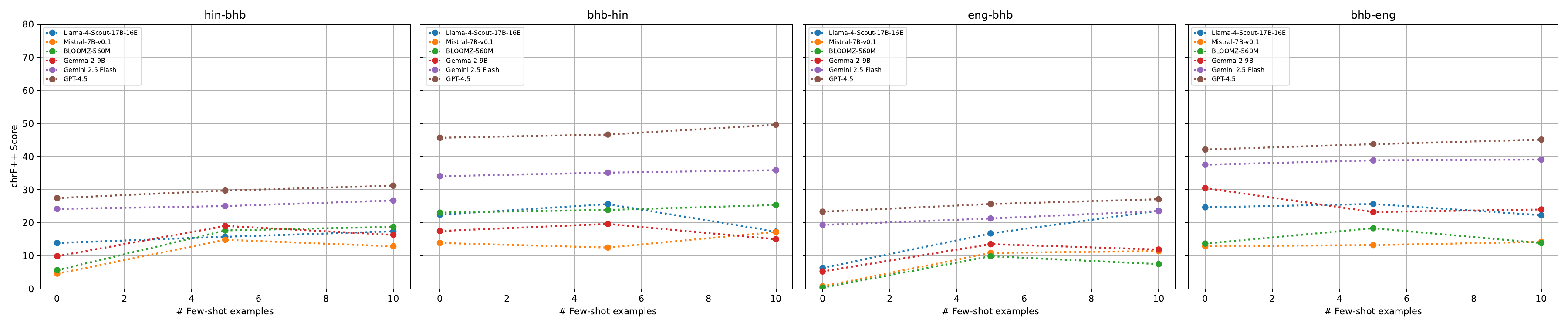}
    \caption{chrF++ performance trends of LLMs across in-context examples (0, 5, 10) shots for four translation directions: Hindi$\leftrightarrow$Bhili and English$\leftrightarrow$Bhili direction. }
    \label{fig:curves models}
\end{figure*}

\section{Experimental Results \& Analysis }

\subsection{Baseline Models}

Despite the proliferation of multilingual language models, significant gaps persist for underrepresented and endangered languages~\cite{protasov2024super,costa2022no}. Even state-of-the-art multilingual transformers such as mT5, NLLB-200, XLM-R~\cite{conneau2020unsupervisedcrosslingualrepresentationlearning}, and decoder-based architectures such as BLOOMZ often underperform on low-resource languages, focusing primarily on well-represented ones.
To establish strong benchmarks for Bhili translation, we leverage the Bhili-Hindi-English Parallel Corpus (BHEPC) and evaluate a range of open-source \& proprietary language models, including
IndicTrans2, NLLB 200, mT5~\cite{xue-etal-2021-mt5}, Qwen3-8B~\cite{yang2025qwen3technicalreport}, DeepSeek-V3~\cite{deepseekai2025deepseekv3technicalreport}, Gemma-2-9B~\cite{gemmateam2024gemmaopenmodelsbased}, Mistral-7B-v0.1~\cite{jiang2023mistral}, BLOOMZ-7B1~\cite{bloomz}, Llama-2-7B~\cite{touvron2023ll}, Llama-3-8B~\cite{grattafiori2024llama3}, Llama-4-Scout-17B-16E~\cite{meta2025llama4scout}, Gemini 2.0 Flash~\cite{googlecloud2025gemini2}, Gemini 2.5 Flash~\cite{googlecloud2025gemini}, GPT-3.5 Turbo~\cite{openai2023gpt35turbo}, GPT-4o-0513~\cite{openai2024gpt4technicalreport}, and GPT-4.5~\cite{openai2025gpt45}  across various model sizes. 
Models are assessed under diverse paradigms, including in-context learning (0,5 and 10 shots), fine-tuning, and cross-domain generalization.

\subsection{In-context learning on BHEPC}

\begin{table*}[!htbp]
\centering
\setlength{\tabcolsep}{5pt} % shrink column spacing
\renewcommand{\arraystretch}{1.1} % row height
\begin{tabular}{lcccccccc}
\toprule
\textbf{Model (LLM)} & \multicolumn{2}{c}{\textbf{hin-bhb}} & \multicolumn{2}{c}{\textbf{bhb-hin}} & \multicolumn{2}{c}{\textbf{eng-bhb}} & \multicolumn{2}{c}{\textbf{bhb-eng}} \\
\cmidrule(lr){2-3} \cmidrule(lr){4-5} \cmidrule(lr){6-7} \cmidrule(lr){8-9}
Eval. Metric & spBLEU & chrF++ & spBLEU & chrF++ & spBLEU & chrF++ & spBLEU & chrF++ \\
\midrule
IndicTrans2 & 9.29 & 35.67 & 27.21 & 53.12 & 5.21 & 31.45 & 15.32 & 42.00 \\
NLLB-200 (600M)   & 11.30 & 42.27 & \colorbox{cyan!30}{\textbf{37.59}} & \colorbox{cyan!30}{\textbf{60.62}} & \colorbox{cyan!30}{\textbf{7.85}} & \colorbox{cyan!30}{\textbf{35.18}} & \colorbox{cyan!30}{\textbf{27.00}} & \colorbox{cyan!30}{\textbf{53.00}} \\
mT5 small   & 11.00 & 42.66 & 29.64 & 54.54 & 5.50 & 27.82 & 16.29 & 41.34 \\
mT5 base    & \colorbox{cyan!30}{\textbf{11.68}} & \colorbox{cyan!30}{\textbf{42.83}} & 34.67 & 58.67 & 7.30 & 33.63 & 19.76 & 45.45 \\
Llama-3-8B  & 10.78 & 27.89 & 19.35 & 35.21 & 4.32 & 21.23 & 11.70 & 30.59 \\
BLOOMZ-7B1  & 9.34 & 23.67 & 14.32 & 32.56 & 3.25 & 19.43 & 8.76 & 29.21 \\
Mixtral-7B-v0.1  & 7.97 & 21.30 & 12.30 & 27.89 & 1.80 & 13.50 & 3.56 & 21.65 \\
Gemma-2-9B   & 8.34 & 25.45 & 13.21 & 29.45 & 3.36 & 17.89 & 7.87 & 28.67 \\
DeepSeek-V3  & 3.56 & 15.34 & 6.12 & 18.46 & 3.87 & 13.23 & 7.45 & 16.34 \\
Qwen3-8B     & 2.12 & 18.79 & 4.86 & 17.96 & 1.32 & 19.42 & 9.97 & 22.27 \\
\bottomrule
\end{tabular}
\caption{spBLEU and chrF++ scores for fine-tuned LLMs. Best scores are highlighted in cyan, with NLLB-200 and mT5-base leading across directions.}
\label{tab:fine}
\end{table*}

Large language models (LLMs) exhibit strong few-shot learning capabilities, effectively performing tasks by leveraging a limited number of exemplars. However, their generative capabilities remain inadequate for underrepresented languages due to imbalances in pretraining data.
In this section, we empirically examine the impact of varying in-context examples on LLM translation performance across Hindi (hin), Bhili (bhb), and English (eng). Experiments with 0, 5, and 10-shot prompting were conducted across all translation directions, and chrF++ scores are reported in Table~\ref{tab:shot}. We benchmarked both open-source and proprietary models, including Mistral-7B-v01, Gemma-2-9B, Llama-2-7B, Llama-3.2-1B, Llama-3-8B, Llama-4-Scout-17B-16E, BLOOMZ (560M, 3.1B, 7B1), Gemini 2.0/2.5 Flash, GPT-3.5 Turbo, GPT-4o, and GPT-4.5 over the BHEPC dataset.

Results indicate that proprietary models consistently outperform open-source models across all directions, with improvements correlated to larger model sizes and more in-context examples (Figure.~\ref{fig:three_images}).
Notably, in Figure.~\ref{fig:curves models} translations from Bhili to English and Bhili to Hindi show sharper gains compared to translations into Bhili. While open-source models show significant improvement from 0 to 5 shots, but gains from 5 to 10 shots are marginal, particularly in hin to bhb and eng to bhb directions.

Among open-source models, Llama-4-Scout-17B-16E performs best for eng→bhb across all shots and for hin→bhb at 0 shots. Gemma leads at 5 shots, and BLOOMZ-560M shows better results at 10 shots. However, Mistral-7B and smaller Llama variants perform poorly across directions. In bhb→hin, BLOOMZ 3.1B and 7B outperform others, while bhb→eng shows degraded performance for Gemma and BLOOMZ-560M beyond 5 shots.
Overall, no consistent trend emerges across models, but chrF++ scores are higher when Bhili is the source language. This is most evident in bhb to eng translations, where Gemma achieves high scores (e.g., 30.50 at 0 shots), likely benefiting from extensive English pretraining. In contrast, translations into Bhili remain challenging, with Llama-4-Scout scoring only 6.35 at 0 shots and other models scoring even lower for eng to bhili direction.

These results highlight that translations into high-resource languages are handled more effectively than translations into low-resource languages like Bhili. This disparity reflects the complex linguistic structure of Bhili and the lack of sufficient resources. Larger models demonstrate greater robustness and benefit more from additional context, but translation quality remains highly dependent on language pair, translation direction, data availability, and model architecture.

% For a detailed contextual analysis, refer to Appendix~\ref{Appendix:8}.

 \subsection{Fine-tuning LLMs on BHEPC
 \& Comparison with In-Context
 Learning}\label{sec: finetune}

% \begin{table*}[!t]
% \centering
% \renewcommand{\arraystretch}{0.98} % Reduced row gap
% \setlength{\tabcolsep}{6pt}
% \begin{tabular}{lccccccc}
% \toprule
% \textbf{Finetuning Corpus} & \textbf{Size} & \multicolumn{5}{c}{\textbf{Testing Corpus}} \\ 
% \cmidrule(lr){3-8}
% & & NCERT & Gov/PMI & Mass Media & NCERT & Gov/PMI & Mass Media  \\ 
% \cmidrule(lr){3-5} \cmidrule(lr){6-8}
% & & \multicolumn{3}{c}{\textit{hin-bhb}} & \multicolumn{2}{c}{\textit{bhb-hin}} \\ 
% \midrule
% NCERT & 10k & \textbf{30.38} & 24.39 & 30.95 & \textbf{54.20} & 36.14 & 44.33 \\ 
% Gov/PMI & 34k & 19.44 & \textbf{38.75} & 37.29 & 33.40 & \textbf{60.09} & 58.04 \\ 
% Mass Media & 64k & 20.50 & 34.68 & \textbf{42.08} & 32.26 & 50.98 & \textbf{61.40} \\
% \cmidrule(lr){1-8}
% \cmidrule(lr){3-5} \cmidrule(lr){6-8}
% & & \multicolumn{3}{c}{\textit{eng-bhb}} & \multicolumn{2}{c}{\textit{bhb-eng}} \\ 
% \cmidrule(lr){3-5} \cmidrule(lr){6-8}
% NCERT & 10k & \textbf{87.35} & 34.51 & 24.77 & \textbf{70.37} & 28.51 & 38.18 \\ 
% Gov/PMI & 34k & 17.68 & \textbf{43.65} & 31.16 & 25.22 & \textbf{47.69} & 45.58 \\ 
% Mass Media & 64k & 20.18 & 29.40 & \textbf{37.31} & 25.57 & 51.11 & \textbf{52.35} \\ 
% \bottomrule
% \end{tabular}
% \caption{chrF++ scores of the fine-tuned NLLB-200 distilled (600M) model under cross-domain generalization for Hindi$\leftrightarrow$Bhili and English$\leftrightarrow$Bhili translation directions. Bold values denote the highest performance per column.}
% \label{tab:finetuning_testing}
% \end{table*}

\begin{table*}[!t]
\centering
\renewcommand{\arraystretch}{1.1} % Increased row gap
\setlength{\tabcolsep}{6pt}
\begin{tabular}{lccccccc}
\toprule
\textbf{Finetuning Corpus} & \textbf{Size} & \multicolumn{5}{c}{\textbf{Testing Corpus}} \\ 
\cmidrule(lr){3-8}
& & NCERT & Gov/PMI & Mass Media & NCERT & Gov/PMI & Mass Media  \\ 
\cmidrule(lr){3-5} \cmidrule(lr){6-8}
& & \multicolumn{3}{c}{\textit{hin-bhb}} & \multicolumn{2}{c}{\textit{bhb-hin}} \\ 
\midrule
NCERT & 10k & \textbf{30.38} & 24.39 & 30.95 & \textbf{54.20} & 36.14 & 44.33 \\ 
Gov/PMI & 34k & 19.44 & \textbf{38.75} & 37.29 & 33.40 & \textbf{60.09} & 58.04 \\ 
Mass Media & 64k & 20.50 & 34.68 & \textbf{42.08} & 32.26 & 50.98 & \textbf{61.40} \\
\cmidrule(lr){1-8}
\cmidrule(lr){3-5} \cmidrule(lr){6-8}
& & \multicolumn{3}{c}{\textit{eng-bhb}} & \multicolumn{2}{c}{\textit{bhb-eng}} \\ 
\cmidrule(lr){3-5} \cmidrule(lr){6-8}
NCERT & 10k & \textbf{87.35} & 34.51 & 24.77 & \textbf{70.37} & 28.51 & 38.18 \\ 
Gov/PMI & 34k & 17.68 & \textbf{43.65} & 31.16 & 25.22 & \textbf{47.69} & 45.58 \\ 
Mass Media & 64k & 20.18 & 29.40 & \textbf{37.31} & 25.57 & 51.11 & \textbf{52.35} \\ 
\bottomrule
\end{tabular}
\caption{chrF++ scores of the fine-tuned NLLB-200 distilled (600M) model under cross-domain generalization for Hindi$\leftrightarrow$Bhili and English$\leftrightarrow$Bhili translation directions. Bold values denote the highest performance per column.}
\label{tab:finetuning_testing}
\end{table*}

In this section, we evaluated the performance of fine-tuned large language models (LLMs) on the BHEPC dataset. Except for IndicTrans2, which was pre-trained exclusively in 22 Indian languages, the majority of the models underwent intensive multilingual pretraining encompassing 100–200 languages. However, none of these models have been pre-trained on the Bhili language that we explore in this work.

Table~\ref{tab:fine} reports fine-tuning results across all translation directions. Each model is fine-tuned on the training data, with early stopping based on the validation set, and performance is reported on the test set. The NLLB-200 distilled 600M variant consistently outperforms other models, achieving the highest chrF++ scores, followed by mT5 small and base variants, which also demonstrate strong generalization.
In contrast, IndicTrans2 performs poorly across all directions, most likely as a result of its specialization and subsequent overfitting on its pre-trained set of 22 Indian languages,  resulting in poor generalization to unseen languages like Bhili. For instance, in the Hindi-to-Bhili translation direction, IndicTrans2 achieves a spBLEU score of 9.29 and a chrF++ score of 35.67 only, highlighting constraints due to its limited pre-training scope. On the other hand, the NLLB 600M and mT5 base models perform well, with spBLEU scores of 11.30 and 11.68 and chrF++ values of 42.27 and 42.83, respectively, suggesting their superior ability to handle the intricacies of this language pair. 

\begin{figure}[!h]
  \centering
  \includegraphics[width=7.5cm, height = 7.5cm]{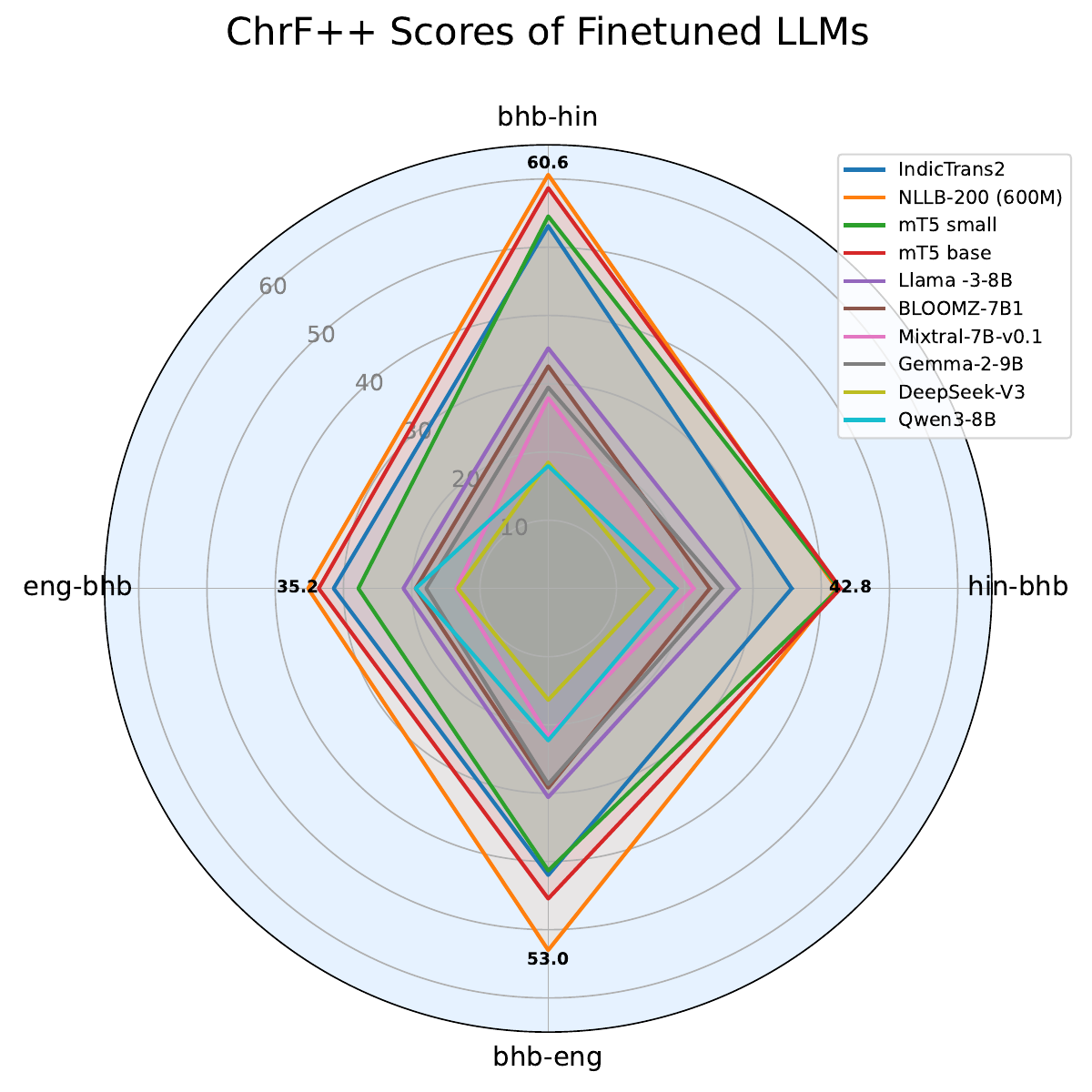}  % fits exactly in one column
  \caption{Radar plot showing chrF++ scores of fine-tuned LLMs across four translation directions. NLLB-200 (600M) excel in low-resource scenarios, while other models exhibit balanced performance.}
  \label{fig:one_column_image}
\end{figure}

In addition, the Mixtral-7B-v0.1, DeepSeek-V3, Qwen3-8B and Gemma-2-9B models  show limited performance, suggesting that increased model capacity alone is insufficient without adequate low-resource language exposure. LLaMA-3-8B and BLOOMZ-7B1 achieve relatively better results, emphasizing the importance of multilingual pretraining quality over sheer model size.

Figure.~\ref{fig:one_column_image} shows chrF++ scores across four translation directions, with each curve representing a model. Models perform better in Bhili$\rightarrow$English and Bhili$\rightarrow$Hindi directions due to richer target language resources, while generating Bhili translations remains challenging. We verified these performance differences using paired bootstrap resampling~\cite{koehn2004statistical}, and report detailed significance tests in Appendix~\ref{app:significance}. Except in the hin→bhb direction, NLLB-200 significantly outperforms all other models (p \textless 0.005).
Comparatively, In-Context Learning (ICL) approaches achieve competitive performance, particularly with larger open-source models like LLaMA-4-Scout-17B-16E, BLOOMZ-7B1, Gemma-9B, and proprietary models such as Gemini 2.5 Flash and GPT-4.5. ICL proves especially effective in low-to-high resource translation directions, benefiting from extensive pretraining and richer contextual examples.

Our findings highlight that fine-tuning remains highly effective for low-resource translation, particularly with models like NLLB-200 and mT5. However, ICL offers a competitive alternative for larger models, reducing the need for expensive fine-tuning while achieving comparable results. This suggests that a hybrid strategy combining fine-tuning for smaller models and ICL for larger ones can effectively optimize translation performance in multilingual, low-resource settings.

%\vspace{-1mm}
\begin{figure*}[!h]
    \centering
    
    %----- Subplot (a)
    \begin{subfigure}[b]{0.24\textwidth}
        \centering
        \includegraphics[trim=60 0 0 0, clip, height=4.2cm]{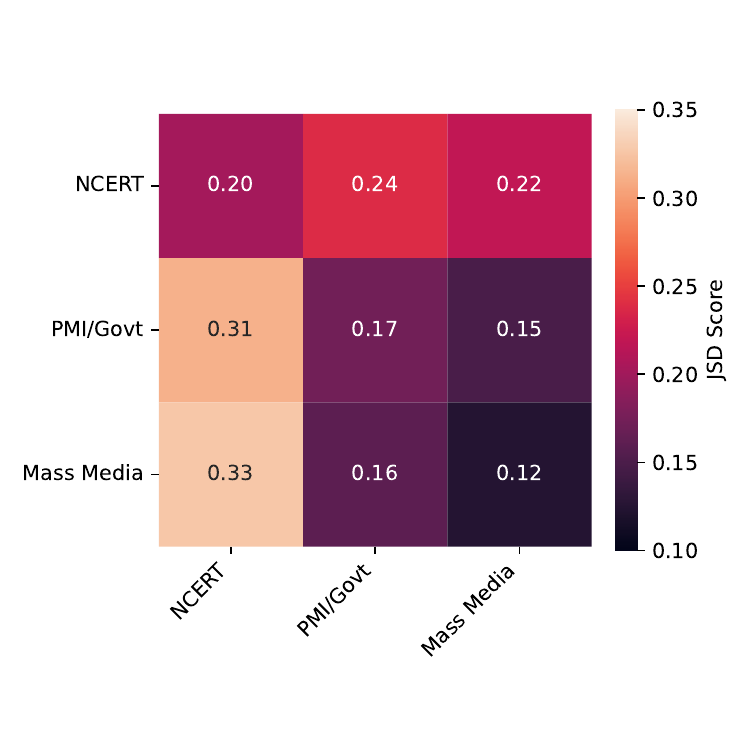} % Replace with your file
        \caption{Hindi to Bhili}
        \label{fig:sub_a}
    \end{subfigure}
    \hfill
    %----- Subplot (b)
    \begin{subfigure}[b]{0.24\textwidth}
        \centering
        \includegraphics[trim=60 0 0 0, clip, height=4.2cm]{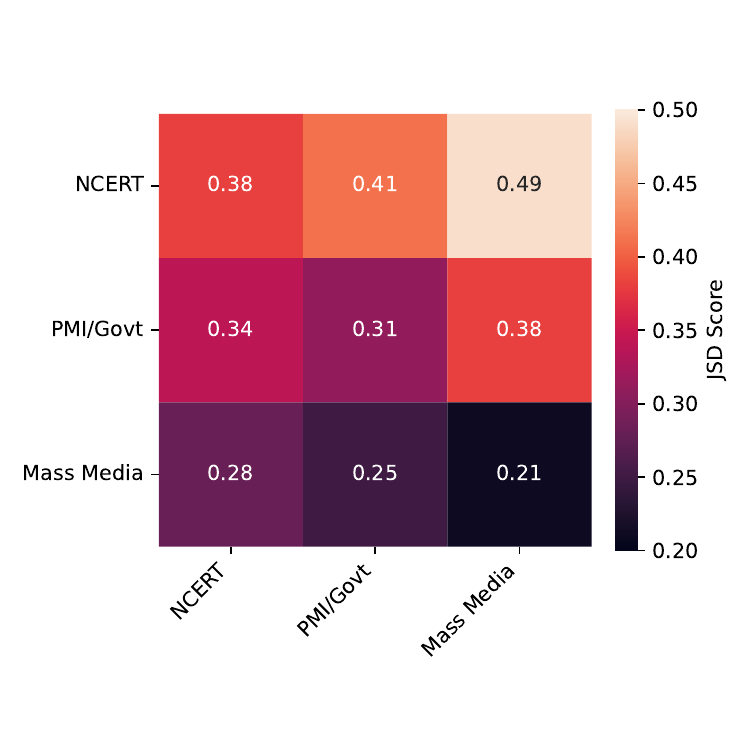} % Replace with your file
        \caption{Bhili to Hindi}
        \label{fig:sub_b}
    \end{subfigure}
    \hfill
    %----- Subplot (c)
    \begin{subfigure}[b]{0.24\textwidth}
        \centering
        \includegraphics[trim=60 0 0 0, clip, height=4.2cm]{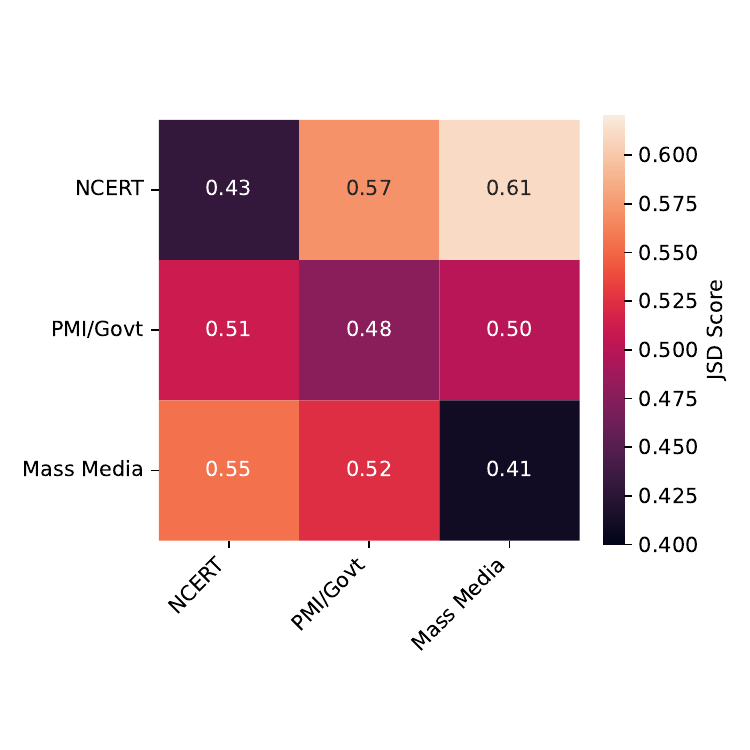} % Replace with your file
        \caption{English to Bhili}
        \label{fig:sub_c}
    \end{subfigure}
    \hfill
    %----- Subplot (d)
    \begin{subfigure}[b]{0.24\textwidth}
        \centering
        \includegraphics[trim=60 0 0 0, clip, height=4.2cm]{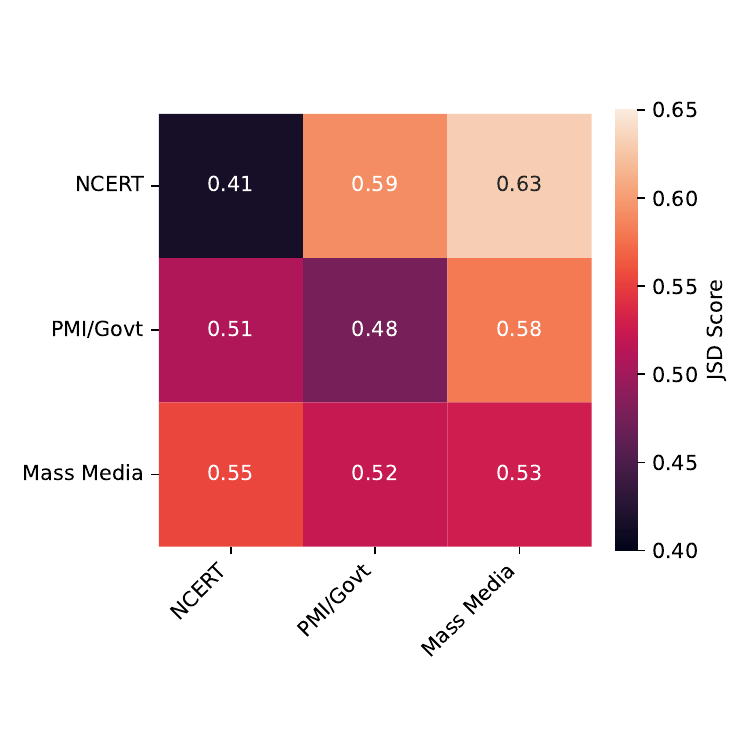} % 
        \caption{Bhili to English}
        \label{fig:sub_d}
    \end{subfigure}
    
    \caption{Jensen-Shannon Divergence (JSD) heatmap for cross-domain generalization evaluation. JSD is computed between in-domain and cross-domain data to quantify distributional divergence between fine-tuning and testing corpora across four translation directions. The results demonstrate that domain shifts significantly impact translation performance, affecting model generalization.}
    \label{fig:four_plots}
\end{figure*}

\subsection{Assessing Cross-Domain Generalization in Machine Translation}
% %\vspace{1mm}

% %\vspace{-2mm}

Domain adaptation remains a critical yet often underexplored challenge in MT for low-resource languages (LRLs). Prior studies show that performance significantly degrades when language models encounter unfamiliar vocabulary and writing styles~\cite{blitzer2008domain,elsahar2019annotate}.
In this section, we investigate two key factors influencing MT performance: (1) the domain similarity between the fine-tuning and testing corpora, and (2) the effect of domain divergence on translation quality.
To analyze cross-domain generalization, we consider three distinct fine-tuning corpora: NCERT textbooks (educational domain), Govt/PMI (administrative/speeches), and Mass Media (News). Performance is assessed across four translation directions:  Hindi$\leftrightarrow$Bhili and English$\leftrightarrow$Bhili. We define in-domain experiments as cases where the fine-tuning and testing corpora originate from the same domain, while cross-domain experiments occur when the domains differ.

As shown in Table \ref{tab:finetuning_testing} in-domain fine-tuning consistently yields higher chrF++ scores compared to cross-domain settings, reinforcing the negative impact of domain shift on translation accuracy. For instance, in the Bhili to Hindi direction, the model fine-tuned on Govt/PMI achieves 60.09 chrF++ scores on the same domain but drops significantly to 33.40 when evaluated on NCERT, 
highlighting
a severe loss in translation quality when faced with cross-domain data.
We quantify domain similarity using Jensen-Shannon Divergence (JSD)~\cite{menendez1997jensen}. The heatmap in Figure.~\ref{fig:four_plots} visualizes JSD scores across different training and testing domain pairs. Darker cells represent lower JSD values, indicating higher domain similarity, while lighter cells signify greater divergence. The results reveal that models fine-tuned on domains with lower JSD (higher similarity) achieve better transaltion quality, whereas those trained on highly divergent corpora struggle to adapt.
These findings emphasize the necessity of domain-aware training for building robust, cross-domain generalizable MT models for low-resource language pairs like Bhili.
For detailed JSD computations and additional analysis, refer to Appendix~\ref{Appendix 5} and~\ref{Appendix 6}.

% By correlating chrF++ and spBLEU scores with JSD, we confirm that domain-relevant fine-tuning data substantially improves generalization and mitigates performance degradation.

% We examined the application of large language models (LLMs) for neural machine translation (NMT) in Bhili, an extremely low-resource language. Our analysis focuses on fine-tuning large-scale pre-trained LLM models on limited parallel data, demonstrating notable improvements in translation quality while identifying persistent challenges.  

% \section{ Qualitative Analysis}

% To assess model shortcomings, we manually analyze the outputs from the fine-tuned NLLB-200 distilled 600M variant, the best-performing model in our experiments. We randomly select 20 sentences from three domains: education, administration, and mass media across four translation directions and got it reviewed by the native speakers, fluent in both Hindi and Bhili, with 1–20 years of linguistic experience.
% We observed several notable error patterns as shown in Figure~\ref{fig:Hindibhb} \&~\ref{fig:engbhb}. Due to Bhili's 70\% lexical similarity with Gujarati, the model frequently produces mixed-language outputs containing Gujarati words or improper verb inflections. Second, hallucination and omission concerns occur in almost all translation directions when the model inserts superfluous content or omits critical information from the source text. Finally, polysemy remains an issue since the model frequently misinterprets English words with various meanings, resulting in inaccurate translations.

\section{Human Evaluation and Qualitative Analysis} \label{sec:human_eval}

Evaluating MT models for low-resource languages like Bhili necessitates a comprehensive assessment framework that combines both quantitative human judgments and qualitative error analysis. This dual perspective ensures a deeper understanding of translation quality beyond surface-level automatic metrics, addressing the linguistic and cultural complexities inherent in low-resource settings.

\subsection{Quantitative Human Evaluation: Alignment with Automatic Metrics} \label{sec:quant_eval}

To systematically assess the correlation between automatic metrics and human judgments of the translation quality,  we conducted a large-scale annotation study following the Multidimensional Quality Metric (MQM)~\cite{sai-b-etal-2023-indicmt, lommel2014multidimensional} 
and Direct Assessment (DA) guidelines. Candidate translations were generated by eight state-of-the-art multilingual models such as IndicTrans2, NLLB-200, mT5-base, Llama-3-8B, BLOOMZ-7B1, Gemma-2-9B, Mixtral-7B-v0.1, and DeepSeek-V3 across four translation directions: Hindi$\leftrightarrow$Bhili and English$\leftrightarrow$Bhili where the segments drawn
from the test set.

After all eight models translated each segment, these source translation pairs
were presented to language experts in randomized
order without revealing the model identities.
For each translation direction, we selected 250 segments and employed two bilingual experts, each a native speaker of the target language and fluent in Hindi and English, to perform evaluations. Annotators highlighted error spans, assigned error categories and severity ratings, and provided DA scores on a 1-5 scale. To ensure annotation consistency, experts initially evaluated 50 shared segments, resolving minor disagreements through discussion. This process yielded a total of 1,000 annotated segments, forming a robust foundation for metric evaluation.
These human-curated annotations subsequently served as test data for spBLEU and chrF++ metric evaluations and underpinned our MQM score computations. The final Inter-Annotator Agreement (IAA) coefficients were 0.60 for Hindi→Bhili, 0.66 for Bhili→Hindi, 0.53 for English→Bhili, and 0.57 for Bhili→English directions, confirming high annotation reliability across language pairs.
In addition, we conducted an inter-annotator agreement (IAA) study on our manually curated Hindi$\rightarrow$Bhili gold references data: two translators independently rendered 250 sentences into Bhili and rated each other’s outputs on the MQM scale, yielding an IAA of 0.59. Since only the Hindi$\rightarrow$Bhili side was human-translated (with English generated via IndicTrans2), we report gold-data IAA only for this direction.
Table~\ref{tab:bhili-proxy-correlations} presents the segment-level Kendall’s $\tau$ and Pearson’s $\rho$ correlations between spBLEU, chrF++, and human MQM scores. The results demonstrate that chrF++ consistently exhibits stronger alignment with human judgments across all translation directions. Notably, the correlation is weakest when Bhili is the target language (hin$\rightarrow$bhb: $\tau$ = 0.18, $\rho$ = 0.25; eng$\rightarrow$bhb: $\tau$ = 0.14, $\rho$ = 0.20), reflecting the challenges models face in accurately generating Bhili translations. 
These errors often involve critical lexical mistranslations and cultural inaccuracies, which receive the most severe MQM penalties. Conversely, when Bhili serves as the source language (bhb$\rightarrow$hin: $\tau$ = 0.28, $\rho$ = 0.36; bhb$\rightarrow$eng: $\tau$ = 0.24, $\rho$ = 0.31), models exhibit fewer critical lexical errors, leading to higher metric correlations and stronger alignment with human adequacy and fluency assessments.
 
% %\vspace{-1.0 mm}

\subsection{Qualitative Error Analysis: Linguistic and Cultural Insights} \label{sec:error}

To complement the quantitative analysis, we conducted a detailed qualitative error analysis focusing on the fine-tuned NLLB-200 distilled 600M variant, identified as the best-performing model in our experiments. We randomly selected 100 sentences from four translation directions. Native speakers of Bhili, with 1 to 20 years of linguistic expertise, reviewed these translations to identify and categorize prevalent error patterns.

Key issues observed include:
\setlength{\itemsep}{-2pt}
\begin{itemize}
\item \textbf{Language Mixing:} Due to Bhili's high lexical overlap with Gujarati, the model frequently introduced Gujarati words or inappropriate verb inflections, particularly in administrative and mass media domains.
\vspace{-2mm}
\item \textbf{Hallucination and Omission:} Across all directions, the model exhibited a tendency to hallucinate content not present in the source or omit essential information, severely affecting translation fidelity.
\vspace{-2mm}
\item \textbf{Polysemy and Lexical Ambiguity:} The model frequently misinterpreted words with multiple meanings, particularly in English-to-Bhili translations, leading to contextually inappropriate outputs.
\vspace{-2mm}
\item \textbf{Domain-Specific Translation Failures:} The model struggled with specialized terminology and formal registers, producing inconsistent translations in the education and administrative domains.
\end{itemize}

Representative examples of these errors are visualized in Figures.~\ref{fig:Hindibhbeorr} and~\ref{fig:engbhb}. Such qualitative insights underscore the need for culturally grounded gold standard datasets and fine-tuning strategies that better capture the linguistic richness and structural characteristics of Bhili.

\section{Conclusion}

% %\vspace{-1mm}
In this work, we present the first large-scale Bhili-Hindi-English Parallel Corpus (BHEPC) and establish strong benchmarks for Bhili translation. Our data acquisition methodology was both resource-intensive and community-engaged, reflecting the complexities of collecting high-quality linguistic data for an under-documented language. 
To assess whether LLMs trained on high-resource Devanagari-script languages generalize to Bhili, we conducted extensive benchmarking. While Bhili shares the Devanagari script with Hindi, our findings revealed significant performance degradation particularly in generation tasks challenging the assumption of positive transfer.
This observation underscores a pivotal insight script-level similarity alone is insufficient to guarantee semantic transfer or effective representation learning. Bhili’s distinct linguistic and cultural characteristics remain underrepresented in existing models, underscoring the need for dedicated LLMs tailored to low-resource languages like Bhili.

\section*{Limitations \& Future Work}
% %\vspace{-1mm}
Although our study introduces a high-utility parallel corpus and benchmarks for Bhili, it is not without constraints. The scarcity of monolingual Bhili data limited us to supervised fine-tuning, restricting the use of unsupervised or semi-supervised approaches. Similarly, we did not experiment with augmentation techniques such as back-translation or pivot-based transfer, which have proven effective in other low-resource NMT contexts. Furthermore, while the creation of a 110k, sentence parallel corpus is a substantial contribution, the manual effort required raises concerns about scalability to the thousands of other low-resource languages worldwide. To mitigate this, we adopted a hybrid workflow that begins with a modest seed corpus and iteratively expands it through model-assisted generation and native speaker post-editing. This approach reduces reliance on exhaustive manual translation, yet broader generalizability and cross-domain robustness remain open challenges for future work.

% Although our study introduces a high-utility parallel corpus and benchmarks for Bhili, it is not without constraints. The limited availability of monolingual Bhili data confined us to supervised fine-tuning, precluding the application of unsupervised or semi-supervised methodologies. Moreover, we did not explore well-established data augmentation strategies such as back-translation or pivot-based augmentation, both of which have been shown to enhance performance in low-resource NMT settings. These limitations potentially impact the generalizability and cross-domain robustness of our models.

\section*{Ethical Statements }

We are committed to upholding the highest ethical standards throughout this work. All human translations and annotations were performed by professional language experts with verified proficiency in the target languages and relevant domain expertise. Annotators were fairly compensated with competitive monthly remuneration, aligned with prevailing government standards and reflective of their linguistic skills and the time and effort invested. Additionally, we utilized publicly available Hindi datasets from the BPCC corpus, released under the CC-0 and CC-BY-4.0 licenses.
All external resources were used strictly in accordance with their intended research purposes, and the resulting BHEPC dataset is intended solely for academic research and not for commercial purposes.
 We have obtained the  consent from all the language experts.

\section*{Acknowledgements}

% We gratefully acknowledge the support of the Ministry of Tribal Affairs, Government of India, and in particular Shri Vibhu Nayar, Secretary, whose encouragement and visionary leadership, together with the generous support of the Adi Vaani Project and the Tribal Research Institute, Bhopal, enabled us to recruit essential human resources and procure the cloud infrastructure required for this work, as well as engage expert human translators for dataset creation. We sincerely thank Smt. Sarika Dhoulpuria,  Mr. Kumar Govind and Mr. Mithun Chandravanshi, for their invaluable assistance in coordinating the data collection process. Most importantly, we express our deep gratitude to all the translators whose dedicated efforts made the construction of the BHEPC benchmark possible. We also extend our thanks to Prof. Parag Singla, Prof. Vivek Kumar, and Dr. Vipul Rathore for their insightful discussions and guidance throughout this project.

We gratefully acknowledge the support of the Ministry of Tribal Affairs, Government of India, and in particular Shri Vibhu Nayar, Secretary, whose encouragement, visionary leadership, and pivotal guidance in shaping the research direction have been essential to advancing this work. Shri Nayar’s foresight in conceptualizing and initiating the Adi Vaani Project has laid the foundation for the preservation and technological empowerment of India’s tribal languages.
His ongoing support and proactive efforts have facilitated the allocation of critical resources and helped create an enabling ecosystem for this research. 
The generous support of the Adi Vaani Project, along with the Tribal Research Institute, Bhopal, has enabled us to recruit essential human resources, procure the cloud infrastructure required for large-scale computational work, and engage expert human translators for dataset creation.
We sincerely thank all officials from the Ministry of Tribal Affairs and the Tribal Research Institute, Bhopal, for their assistance in coordinating data collection. We also express our deep gratitude to all translators for their dedicated efforts in creating the BHEPC benchmark, and to Prof. Parag Singla, Prof. Vivek Kumar, and Dr. Vipul Rathore for their guidance throughout this project.

\section*{Dataset Availability }

The Bhili-Hindi-English Parallel Corpus (BHEPC) presented in this study constitutes the first large-scale, high-quality resource for the extremely low-resource Bhili language. Curated through community-driven efforts, BHEPC aims to facilitate research in low-resource machine translation and promote the digital inclusion of marginalized tribal communities. Given the cultural sensitivity of the Bhili language, its endangered status, and the ongoing research objectives associated with this work, access to the BHEPC dataset will be provided upon request to researchers and institutions for academic and non-commercial purposes.

\bibliography{acl_latex}

\appendix

\section{Appendix}
\label{sec:appendix}

% \begin{table*}[!htbp]
% \centering
% \small
% \setlength\tabcolsep{6pt}
% \begin{tabular}{l|cc|cc|cc|cc}
% \toprule
% \bf Metric    & \multicolumn{2}{c|}{\bf hin→bhb} & \multicolumn{2}{c|}{\bf bhb→hin} & \multicolumn{2}{c|}{\bf eng→bhb} & \multicolumn{2}{c}{\bf bhb→eng} \\
%               & $\tau$ & $\rho$ & $\tau$ & $\rho$ & $\tau$ & $\rho$ & $\tau$ & $\rho$ \\
% \midrule
% spBLEU     & 0.18 & 0.25 & 0.28 & 0.36 & 0.14 & 0.20 & 0.24 & 0.31 \\
% chrF++        & 0.20 & 0.30 & 0.30 & 0.45 & 0.15 & 0.22 & 0.25 & 0.38 \\
% \bottomrule
% \end{tabular}
% \caption{Segment-level Pearson $\rho$ and Kendall’s $\tau$ correlations of spBLEU and chrF++ with human MQM judgments. chrF++ shows stronger alignment with human evaluations across all translation directions.}
% \label{tab:bhili-proxy-correlations}
% \end{table*}

\subsection{Extremely low-resource languages}

Languages with extremely low resources are characterized by a severe scarcity of accessible data and documentation. In the context of Indian regional languages, many fall into this category, where available resources are minimal compared to more widely studied languages. 
Many of these languages are either not published or have very little data available, and they are often said to be under-documented, under-resourced, or under-digitized. Therefore, massive obstacles exist when trying to gather and process raw textual data in these languages.

\subsection{Bhili Language}

Approximately 13 million people across the Indian states of Rajasthan, Gujarat, Maharashtra, and Madhya Pradesh speak Bhili, a Western Indo-Aryan language written in the Devanagari script and deeply rooted in Bhil culture. The dataset we present cover Bhili dialect spoken  in the Madhya Pradesh Jhabua region. Despite its significance, due to the lack of publicly available parallel corpora, Bhili has been mostly unexplored in the domains of NLP and machine translation. Given its large speaker base and close lexical ties to Gujarati and Marathi, developing a robust MT system for Bhili particularly for Hindi-Bhili and English-Bhili language pairs has the potential to bridge critical communication gaps and enhance digital inclusion. 
The growing need for effective digital communication in Bhili-speaking regions emphasizes the potential impact of such a system, making the consolidation of existing resources and the creation of new parallel corpora a critical step toward enabling seamless interaction between Bhili speakers and the broader global community.

\subsection{ Pre-trained Multilingual LLMs }
% change it
Pretrained multilingual models have revolutionized the field of natural language processing (NLP) by making substantial advances in machine translation (MT) and cross-lingual transfer learning.
Despite the growing number of large-scale Multilingual language models  such as IndicTrans2 (supports 22 Indian languages), NLLB (covering 200 languages), and mT5 (spanning 101 languages), Bhili remains largely  overlooked. This highlights a critical gap in multilingual MT frameworks. Similarly, even the latest large language models, such as Gemma, Mixtral, DeepSeek,Qwen3, the Llama , and BLOOMZ family series, have expanded multilingual representation but still do not include Bhili, making it less accessible for computational applications.
Beyond academic research, commercial MT platforms like Google Translate~\footnote{Google Cloud Translation API Reference: \url{https://cloud.google.com/translate/docs/reference/rest}.}
 and Microsoft Translator~\footnote{Microsoft Translator API Reference: \url{https://www.microsoft.com/en-us/translator/business/translator-api/}.}
 also exclude Bhili, limiting its digital presence and practical usability. This lack of representation both in research-driven and commercial models makes it even harder to preserve the language, improving accessibility, and integrating Bhili into modern NLP applications.

% \subsection{Training details: }

%  These studies use the Gemma-2-7B and Gemma-2-27B
% models as baselines, combining zero-shot translation with Low-Rank Adaptation (LoRA) [5] finetuning. Nemotron-Mini-Hindi-4B [7], based on Nemotron-Mini-4B, uses the Supervised Fine-Tuning
% (SFT)

 \subsection{Training Details}

We evaluated a wide range of pre-trained open-source models for low-resource language translation, including both encoder-decoder and decoder-only architectures. 
For all models, we maintain a consistent experimental setup to ensure fair comparison. We explore both full fine-tuning and Parameter-Efficient Fine-Tuning (PEFT) using LoRA (Low-Rank Adaptation)~\cite{hu2022lora}. Hyperparameters are selected through an extensive grid search over batch sizes (8,16, 32) and learning rates (5e-3, 1e-3, 5e-4, 1e-4, 5e-5, 1e-5), and final selections are based on the chrF++ validation scores for both in-domain and cross-domain datasets. The complete hyperparameter configurations are provided in Table~\ref{tab:hyperparams}. In ICL experiments, exemplars are randomly sampled from the training set. For all decoder-based models, we set the decoding temperature to 0.1 to avoid degenerate outputs.
For all translation directions, we applied a uniform prompt across all models, as shown in Table~\ref{tab:prompts}. 
Given the high computational demands of fine-tuning large models, we performed only a single run per fine-tuning experiment rather than averaging results across multiple runs.

\subsection{Computing Infrastructure}
All experiments are performed on a High Performance Computing Cluster having NVIDIA A100 GPUs. Model training times range from 6 to 48 hours, depending on model size and dataset scale. The Hugging Face Transformers library is used for model implementation and fine-tuning, while evaluation metrics are computed using NLTK~\cite{bird-loper-2004-nltk} and SacréBLEU~\cite{post-2018-call}. LoRA experiments leverage the PEFT library for efficient adaptation.

\begin{table*}[!htbp]
\centering
\small
\setlength\tabcolsep{6pt}
\begin{tabular}{l|cc|cc|cc|cc}
\toprule
\bf Metric    & \multicolumn{2}{c|}{\bf hin→bhb} & \multicolumn{2}{c|}{\bf bhb→hin} & \multicolumn{2}{c|}{\bf eng→bhb} & \multicolumn{2}{c}{\bf bhb→eng} \\
              & $\tau$ & $\rho$ & $\tau$ & $\rho$ & $\tau$ & $\rho$ & $\tau$ & $\rho$ \\
\midrule
spBLEU     & 0.18 & 0.25 & 0.28 & 0.36 & 0.14 & 0.20 & 0.24 & 0.31 \\
\rowcolor{green!20}  % Light green highlighting
chrF++        & 0.20 & 0.30 & 0.30 & 0.45 & 0.15 & 0.22 & 0.25 & 0.38 \\
\bottomrule
\end{tabular}
\caption{Segment-level Pearson $\rho$ and Kendall’s $\tau$ correlations of spBLEU and chrF++ with human MQM judgments. chrF++ shows stronger alignment with human evaluations across all translation directions.}
\label{tab:bhili-proxy-correlations}
\end{table*}

\subsection{Jensen-Shannon Divergence (JSD) for Cross-Domain Generalization Analysis}\label{Appendix 5}

Jensen-Shannon Divergence (JSD) quantifies the similarity between two probability distributions, \( A \) and \( B \), and is defined as:

\begin{equation}
JSD(A \parallel B) = \frac{1}{2} KL(A \parallel M) + \frac{1}{2} KL(B \parallel M)
\label{1}
\end{equation}

where \( M \) is the mean distribution, and \( KL(\cdot \parallel \cdot) \) represents the Kullback-Leibler divergence.

As shown in Equation~\eqref{1}, the JSD~\cite{menendez1997jensen} is a symmetrized version of the KL divergence. To compute JSD across domains, we tokenize text using
\texttt{bert\-/base\-/multilingual\-/cased}, process batches efficiently,
normalize numeric and temporal expressions, and construct token frequency
distributions. Missing tokens across corpora are assigned zero probability
for proper alignment.
We evaluate JSD for all translation directions across the NCERT, Govt/PMI,
and Mass Media domains. Heatmap visualizations in Figure~\ref{fig:four_plots}
show domain shifts, where lower JSD values indicate higher similarity. These
insights inform cross-domain adaptation strategies in neural machine
translation (NMT), helping mitigate distributional disparities and enhance
model robustness.

 \begin{table*}[!t]
\centering
\small
\renewcommand{\arraystretch}{1.4}
\begin{tabular}{@{}p{5.5cm}|p{5.5cm}@{}}
\toprule
\textbf{Hyperparameters} & \textbf{Values Used} \\
\midrule
Optimizer & Adam \\
Beta Values $(\beta_1, \beta_2)$ & (0.9, 0.98) \\
Learning Rate & 5e-4 \\
Scheduler & Inverse Sqrt \\
Loss Criterion & Cross-Entropy \\

Max Gradient Norm & 1.0 \\
Weight Decay & 0.01 \\
Batch Size & 16 \\
Gradient Accumulation Steps & 4 \\
Patience (Early Stopping) & 10 \\
Mixed Precision Training & FP16 \\
LoRA Rank (r) & 16 (LoRA FT only) \\
LoRA Alpha & 32 (LoRA FT only) \\
LoRA Dropout & 0.1 (LoRA FT only) \\

% Max Training Epochs & 50 \\

Decoding Temperature  & 0.7 \\

\bottomrule
\end{tabular}
\captionsetup{justification=centering, font=small}
\caption{Unified hyperparameter configuration across all models and experiments}
\label{tab:hyperparams}
\end{table*}

\subsection{Cross Domain Impact Analysis on Machine Translation: Correlating spBLEU, chrF++, and JSD Scores}\label{Appendix 6}

To further analyze the impact of domain shift on machine translation performance, we evaluate spBLEU scores across different translation directions and examine the relationship between domain similarity (JSD scores) and translation quality metrics (spBLEU and chrF++).

\begin{figure*}[!t]
    \centering
    % Make sure the image is large enough to cover both columns
    \includegraphics[width=0.99\textwidth]{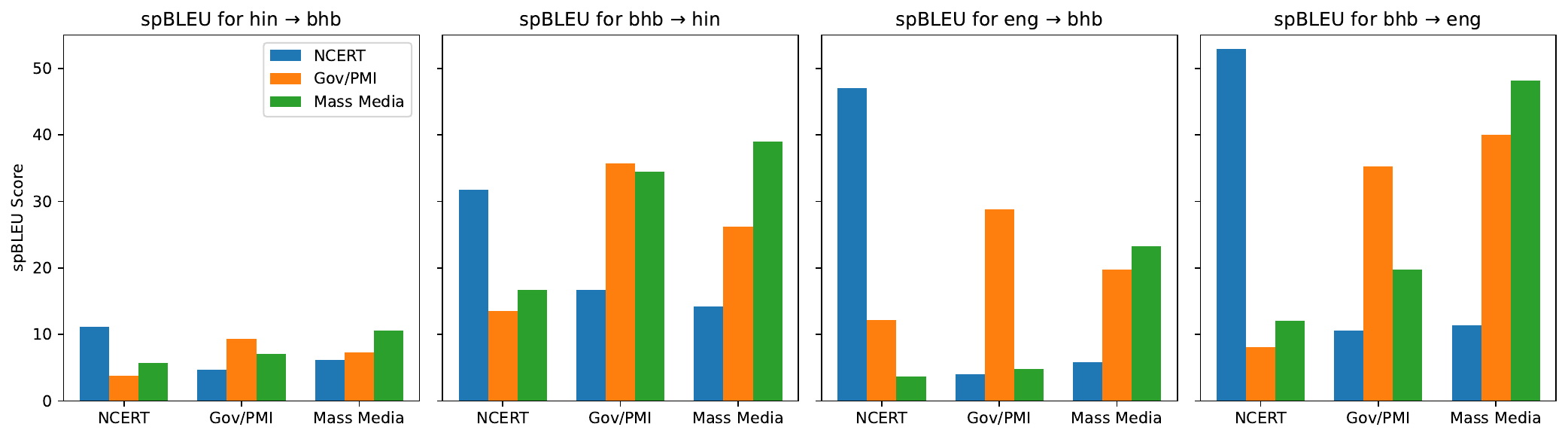}
    \caption{Bar plots showing spBLEU scores across four translation directions:
      (1) \texttt{hin} $\rightarrow$ \texttt{bhb}, (2) \texttt{bhb} $\rightarrow$ \texttt{hin},
      (3) \texttt{eng} $\rightarrow$ \texttt{bhb}, and (4) \texttt{bhb} $\rightarrow$ \texttt{eng}. The  NLLB model fine-tuned on domain-specific datasets: NCERT, Govt/PMI, and Mass Media. The evaluation is conducted on both in-domain and cross-domain data. Each bar represents the translation quality achieved for a given direction and training corpus.}
    \label{fig:spspBLEU_all_directions}
\end{figure*}

Figure.~\ref{fig:spspBLEU_all_directions} presents bar plots of spBLEU scores for four translation directions: Hindi to Bhili, Bhili to Hindi, English to Bhili, and Bhili to English, across three domain-specific fine-tuning settings (NCERT, Govt/PMI, Mass Media). Each bar indicates the translation performance when a model trained on one domain is tested on another. Higher bars signal closer alignment between training and testing domains, while lower bars highlight the effects of domain mismatch. The results indicate that in-domain fine-tuning leads to consistently higher spBLEU scores, whereas cross-domain settings exhibit performance degradation. Notably, in the Bhili to English direction, models fine-tuned on Mass Media outperform those trained on NCERT and Govt/PMI, suggesting that domain alignment plays a crucial role in translation effectiveness.

\begin{figure}[!htbp]
    \centering
    \includegraphics[width=7.6cm, height =5cm]{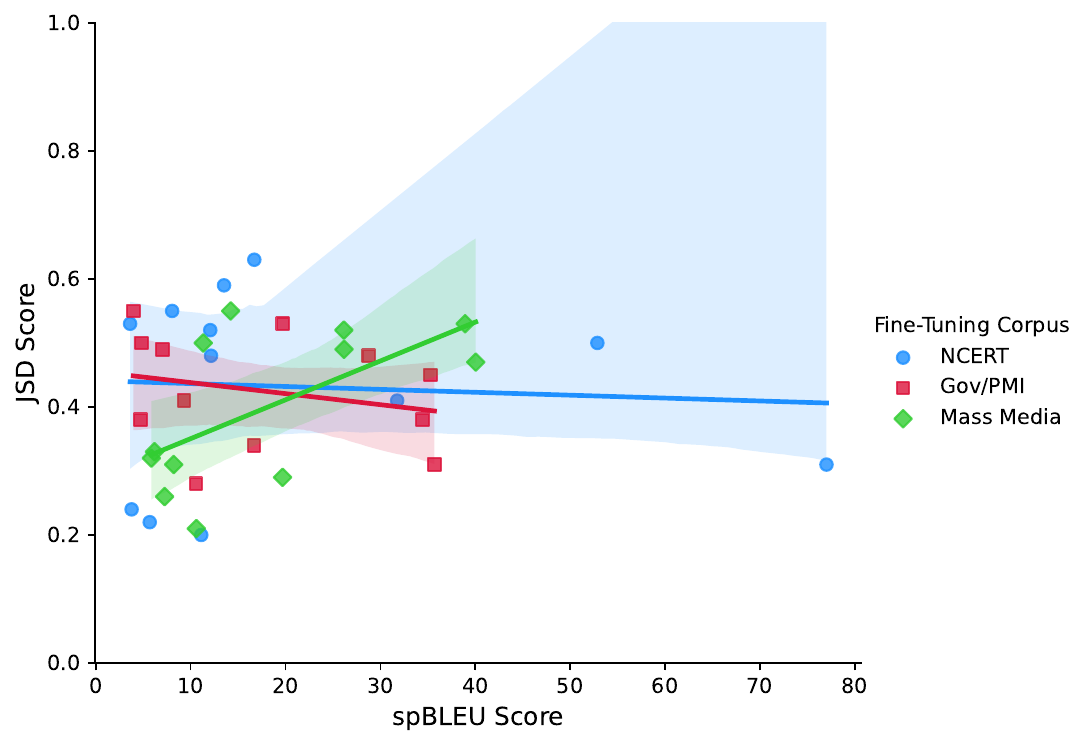}
    \caption{
Plot showing the relationship between JSD and spBLEU scores for the NLLB model across three domains. Data points are domain color-coded, with regression lines and confidence intervals highlighting domain-specific trends. NCERT shows little correlation, while Govt/PMI and Mass Media exhibit slight positive correlations, suggesting a trade-off between JSD and spBLEU scores. }
    \label{fig:diver}
\end{figure}

Figure.~\ref{fig:diver} examines the relationship between Jensen-Shannon Divergence (JSD) and spBLEU scores using scatter plots with regression curves. The plot highlights how domain divergence (JSD) impacts translation quality (spBLEU), with trends varying across fine-tuning corpora. A weak or negative correlation is observed in the NCERT setting, suggesting that higher domain divergence leads to lower translation quality, whereas Govt/PMI and Mass Media show a slight positive correlation. The shaded confidence intervals indicate variability, emphasizing the influence of domain adaptation on model performance. 

% \begin{figure}[t!]
%     \centering
%     \includegraphics[width=0.9\linewidth]{Images/curvesratio (1).pdf}
%     \caption{The spBLEU/JSD ratio across four translation directions (hin→bhb, bhb→hin, eng→bhb, bhb→eng) for the NLLB fine-tuned model on NCERT, Gov/PMI, and Mass Media. Each bar represents the alignment between lexical distribution (JSD) and translation quality (spBLEU), with higher bars indicating a better balance between accuracy and divergence.}
%     \label{fig:spBLEUJSD}
% \end{figure}

\begin{figure*}[ht]
  \centering
  \begin{subfigure}[b]{0.32\textwidth}
    \includegraphics[width=\linewidth]{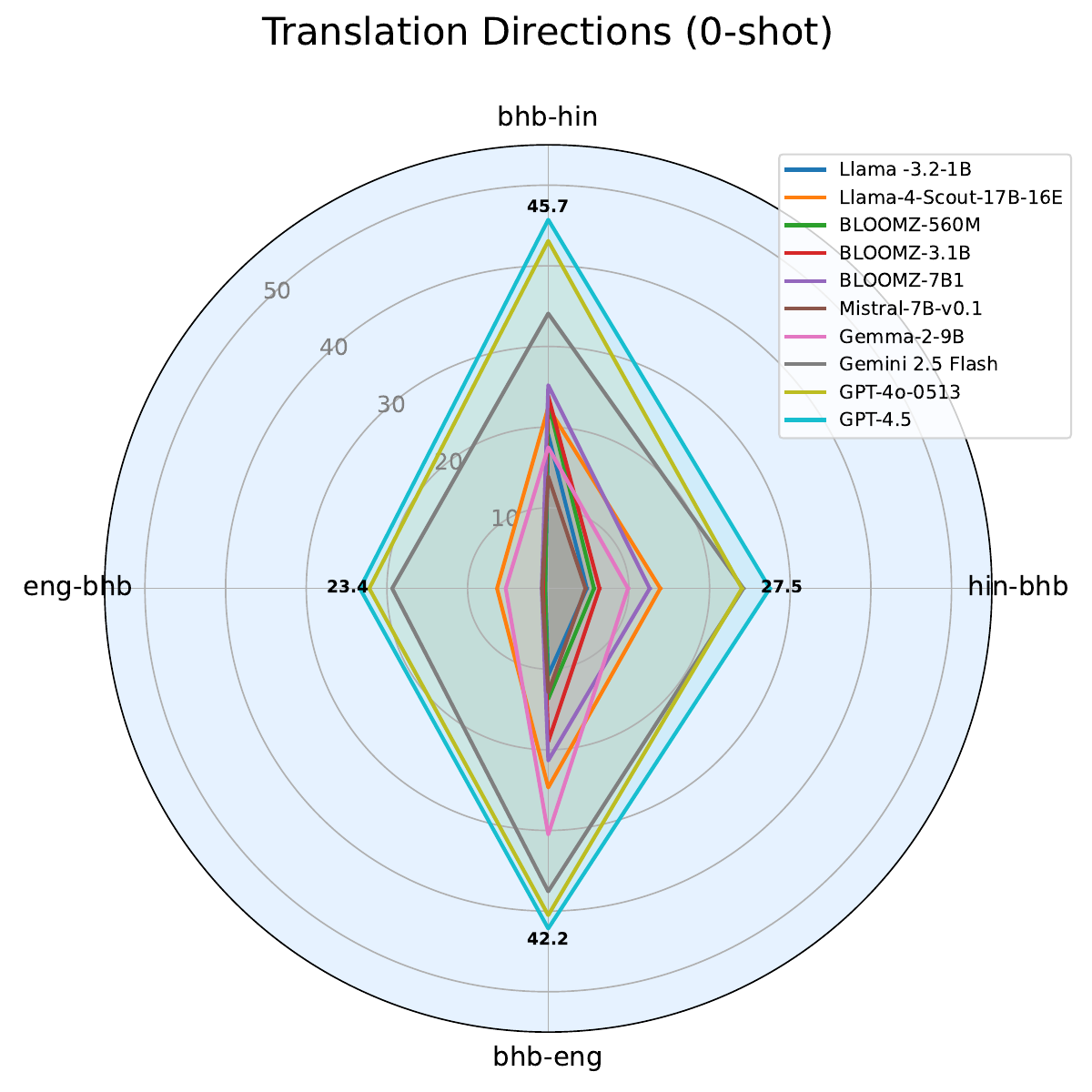}
    \caption{0-shot}
    \label{fig:image1}
  \end{subfigure}
  \hfill
  \begin{subfigure}[b]{0.32\textwidth}
    \includegraphics[width=\linewidth]{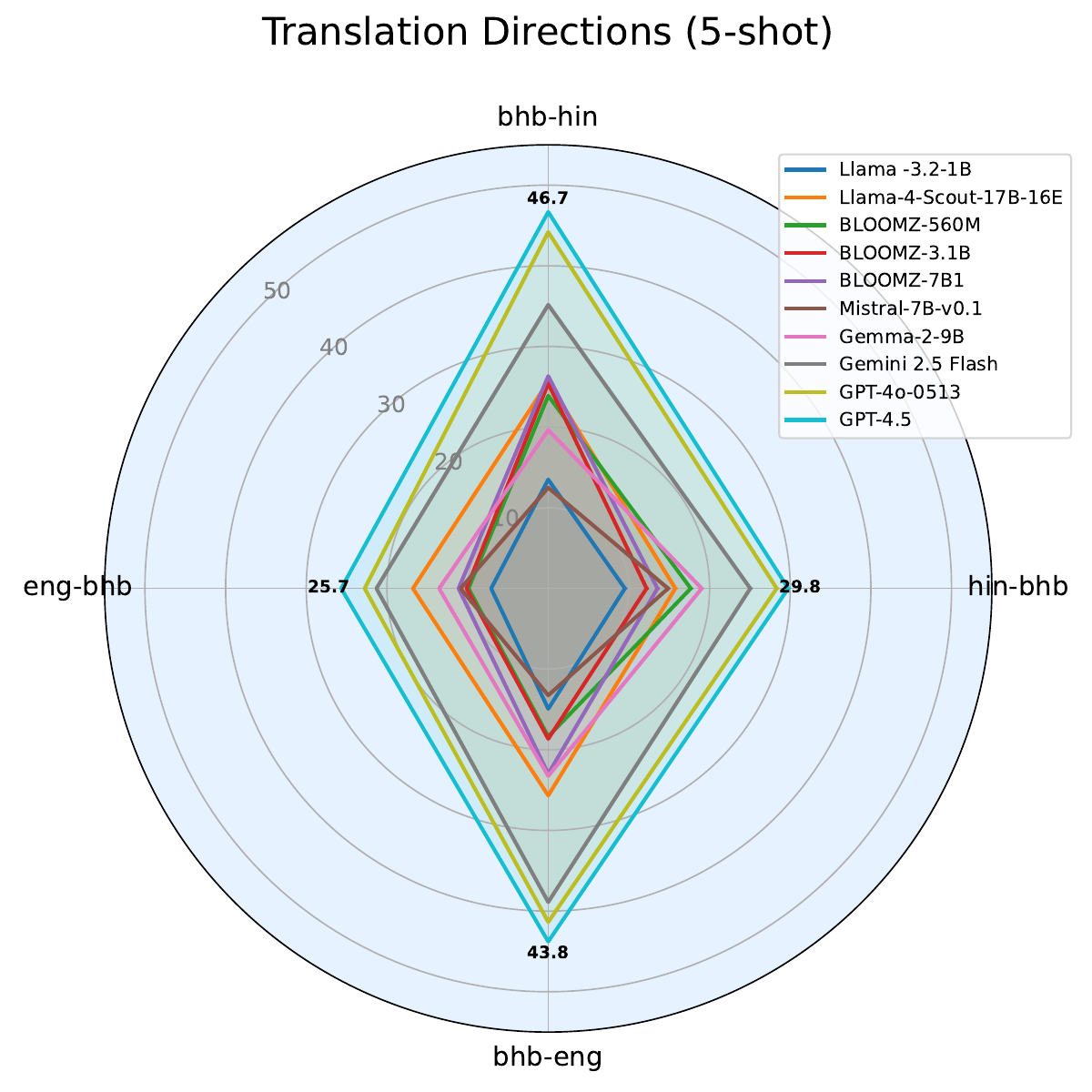}
    \caption{5-shot}
    \label{fig:image2}
  \end{subfigure}
  \hfill
  \begin{subfigure}[b]{0.32\textwidth}
    \includegraphics[width=\linewidth]{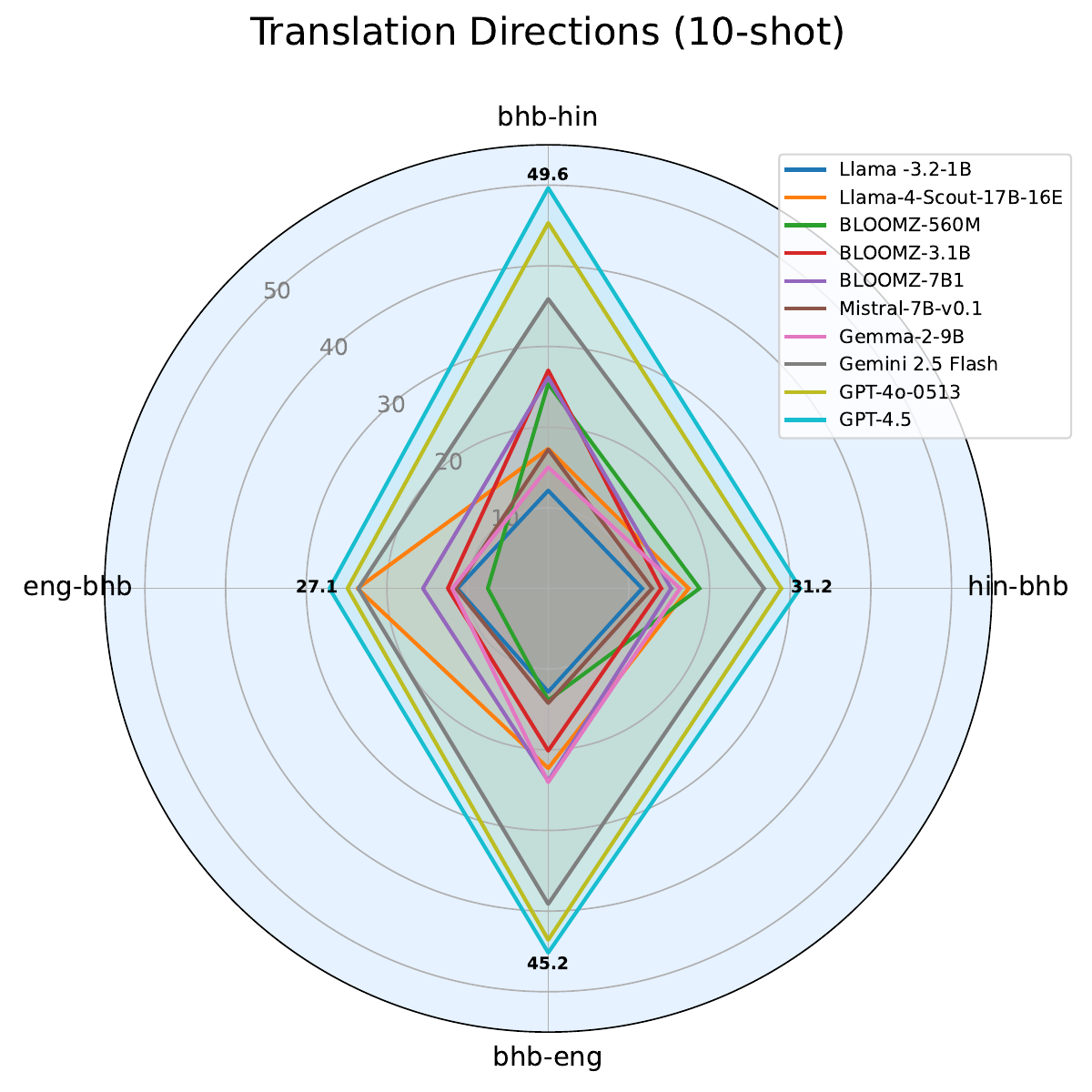}
    \caption{10-shot}
    \label{fig:image3}
  \end{subfigure}
  \caption{Comparison of model performance across four translation directions (hin→bhb, bhb→hin, eng→bhb, bhb→eng) under varying few-shot scenarios. The radar plots highlight that larger models, particularly GPT-4.5 and Llama-4-Scout-17B-16E, consistently outperform smaller models across all settings, with noticeable performance gains as the number of shots increases.}
  \label{fig:three_images}
\end{figure*}

\begin{figure*}[!htbp] % The * makes it span both columns
    \centering
    \includegraphics[width=\textwidth]{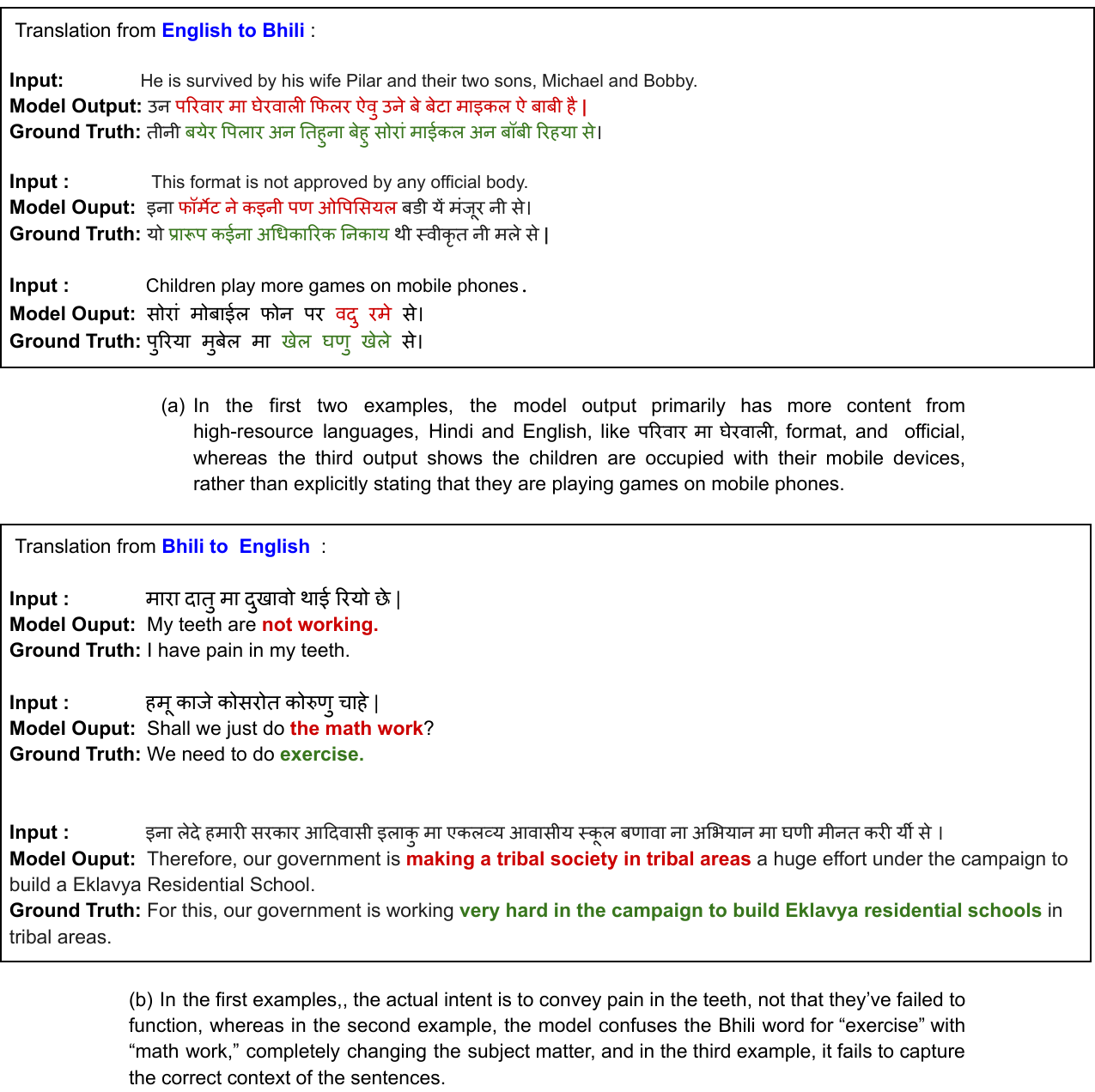}
    % %\vspace{-50pt}% Full width
    \caption{Error analysis in the predictions from the fine-tuned NLLB-200 distilled 600M variant model on the evaluation dataset highlighting errors in the model prediction from English to Bhili and Bhili to English direction.}
    \label{fig:Hindibhbeorr}
\end{figure*}

\begin{figure*}[!htbp] % The * makes it span both columns
    \centering
    \includegraphics[width=\textwidth]{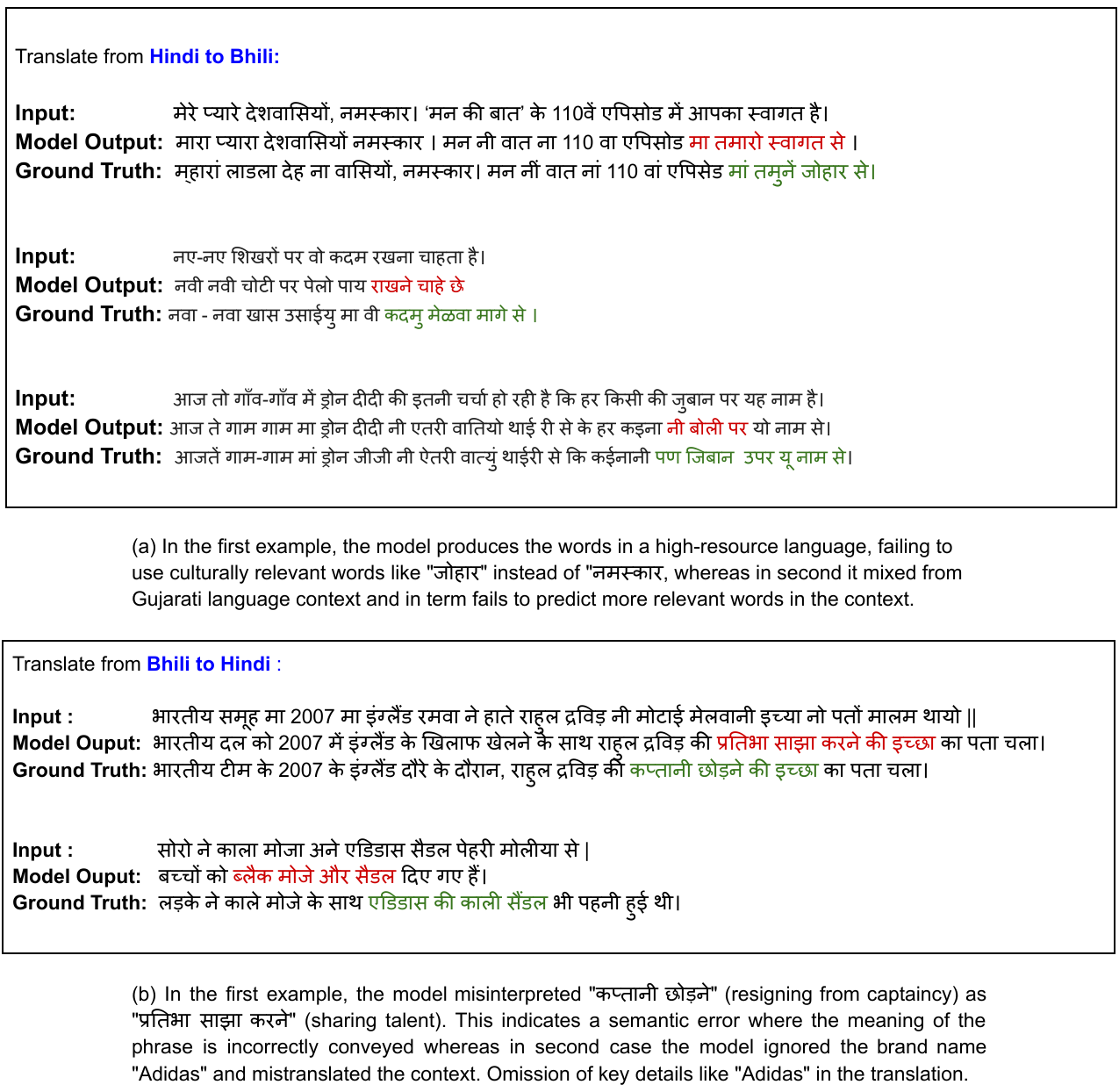}
    % %\vspace{-50pt}% Full width
    \caption{Error analysis in the  predictions from the fine-tuned NLLB-200 distilled 600M variant model on the evaluation dataset highlighting errors in the model prediction from Hindi to Bhili and Bhili to Hindi direction.}
    \label{fig:engbhb}
\end{figure*}

\begin{table*}[!t]
    \centering
    \small % Reduce font size
    \renewcommand{\arraystretch}{2.1} % Adjust row spacing
    \setlength{\tabcolsep}{2.1pt} % Reduce column padding
    \resizebox{\textwidth}{!}{ % Scale to full text width (both columns)
    \begin{tabular}{p{3.5cm} p{11cm}} % Adjust column widths as needed
        \toprule
        \textbf{Translation Direction} & \textbf{Prompt} \\
        \midrule
        
        \textbf{English to Bhili} & Translate the following English sentence to Bhili:\\
        & \textbf{Input:} \textbf{[Sentence in Source Language]} \\ 
        & \textbf{Output:} \\ 
        \midrule
        \textbf{Bhili to English} & Translate the following Bhili sentence to English: \\
        & \textbf{Input:} \textbf{[Sentence in Source Language]} \\ 
        & \textbf{Output:} \\ 
        \midrule
        \textbf{Hindi to Bhili} & Translate the following Hindi sentence to Bhili:\\
        & \textbf{Input:} \textbf{[Sentence in Source Language]} \\ 
        & \textbf{Output:} \\ 
        \midrule
        \textbf{Bhili to Hindi} & Translate the following Bhili sentence to Hindi:\\
        & \textbf{Input:} \textbf{[Sentence in Source Language]} \\ 
        & \textbf{Output:} \\ 
        \bottomrule
    \end{tabular}
    } % End of resizebox
    \caption{Prompt templates used for different translation directions for all the multilingual LLMs. N-shot examples follow the same format as the last test example given to the model.}
    \label{tab:prompts}
\end{table*}

% \begin{figure*}[!htpb]
%     \centering
%     \includegraphics[width=\textwidth]{Images/table-2.pdf} % Full 
%   \caption{Snapshot inserted as a full-width figure in ACL format.}
%     \label{fig:snapshot}
% \end{figure*}
\subsection{Translation Guidelines}

To ensure consistency and semantic fidelity in translation, we developed a comprehensive set of guidelines that balances linguistic rigor with practical limitations (notably the lack of Bhili-specific glossaries, literature and  linguistic resources). Translators are instructed to preserve the source content’s meaning and stylistic register without introducing or omitting content, while handling named entities, numerals, dates, and technical vocabulary in accordance with target language’s conventions. Specifically:

\begin{itemize}

 \item \textbf{General Principles:} Faithfully reproduce the source text’s meaning, tone, style, and register whether formal, colloquial, or emphatic without additions or deletions; correct minor typos or grammatical slips while preserving any factual inconsistencies; and ensure the translation reads fluently and naturally.

 \item \textbf{Named Entities:} Use established conventional translations where available; otherwise, transliterate entities accurately into the target script; and strictly follow language-specific norms without inventing alternative renderings.

 \item \textbf{Numbers \& Units:} Mirror the source’s numeric format exactly (spelled out or in digits); apply local counting conventions for large values while retaining “billion”/ “trillion” in English or accepted local terms; and preserve the original units of measurement.
 
 \item \textbf{Dates:} Maintain the exact date format, whether fully spelled or numeric, and keep the same digit length for years, avoiding any expansion or contraction.

\end{itemize}

\subsection{Annotation Guidelines Based on MQM: Error Categories and Severities}~\label{Appendix:B}

Annotators evaluate translations at the segment level, a segment can consist of a single sentence or multiple sentences by aligning each translated unit with its original source and presenting both sides by side. In Table~\ref{tab:error-hierarchy}, we describe each error type with a clear hierarchy, illustrating how errors are structured across categories. Each category is then assigned a severity rating on a five-point scale (Very low, Low, Medium, High, and Very high) thereby enabling fine-grained distinctions in error impact.
Table~\ref{tab:error-severity} presents the descriptors at the scale’s endpoints as shown to the annotators. To translate their judgments into numeric scores, we assign a weight of 1 for very low, 2 for low, 3 for medium, 4 for high, and 5 for very high. Furthermore, each subcategory such as Accuracy, Fluency, Terminology Inappropriate, and Style, is paired with its own severity marking, so we treat them all with equal significance. Errors unrelated to translation automatically receive a zero score, and any sentence flagged for a source error is omitted from the evaluation.

The following instructions were shared with the annotators:

 \begin{itemize}
   
     \item Annotators were instructed to scrutinize each translated segment and pinpoint every error present, with a strict limit of five errors per segment. Whenever a segment contains more than five mistakes, they should then select and report only the five most consequential errors.

     \item Initially, mark the exact span of text by applying color coding; then select the appropriate category/sub category and assign a severity level from the available options. If the error stems from the original content or represents an omission, the highlighted fragment may instead reside within the source segment to ensure the correct context is captured.

     \item Identify errors at the finest possible granularity. For example, if two words in a sentence are mistranslated, log two separate mistranslation errors.
     
\item In instances where multiple errors overlap within the same text segment, record only the single most severe error; if their severity levels are equal, choose the first matching category in the error typology (e.g., Accuracy, then Fluency, then Terminology).

\item Treat Source error and Non-translation as special cases: annotate Source errors by highlighting the relevant span in the source segment (such sentences are exempt from scoring, though the source error must still be marked).

\item If the translation is so heavily distorted or entirely unrelated that discrete errors cannot be reliably distinguished, flag a single Non-translation error spanning the entire segment no other errors should be noted when this category is selected.

\item Finally, after annotating all errors, assign each translation a score out of 5 and record this value in the final score column.

\end{itemize}

\begin{table*}[!t]
\centering
\small
\renewcommand{\arraystretch}{1.6}
\begin{tabular}{@{}p{3.65cm}|p{11.5cm}@{}}
\toprule
\textbf{Error Category} & \textbf{Explanation} \\
\midrule
\textbf{Accuracy} & \textbf{Addition} \newline
The translation injects information absent from the original source, constituting extraneous content.
\newline
\textbf{Omission} \newline
Translation is missing content from the source. \newline
\textbf{Mistranslation} \newline
The target text fails to faithfully render the semantic intent of the source. \newline
\textbf{Untranslated text} \newline
Source text has been left untranslated \\
\midrule
\textbf{Fluency} & \textbf{Orthographic Inconsistency} \newline
Spelling or capitalization deviates from standard conventions. \newline
\textbf{Syntactic Inaccuracy} \newline
Grammatical constructions are erroneous, excluding orthographic faults. \newline
\textbf{Register} \newline
The level of formality or pronoun usage is contextually inappropriate.
\newline
\textbf{Character Encoding} \newline
Character corruption arises from improper encoding  \\
\midrule
\textbf{Terminology Inappropriate} & Term selection is non-standard or ill-suited to the domain context. \\
\midrule
\textbf{Style Awkward} & The tone or sentence structure is discordant with the genre or unduly verbose.

(Example: 1. The source sentence feels formal like in a newspaper, but the translation doesn’t.

2. Sentences are correct, but simply too long, etc..) \\
\midrule
\textbf{Transliteration} & If it transliterates instead of translating words/phrases, where it should not. \\
\midrule
\textbf{Other} & Any issue not encompassed by the specified categories. \\
\midrule
\textbf{Source Error} & An error residing in the original source that requires annotation. \\
\midrule
\textbf{Non Translation} & The segment is so garbled or unrelated reliably characterize the 5 most severe errors. \\
\bottomrule
\end{tabular}
\captionsetup{justification=centering, font=small}
\caption{Hierarchy of errors accompanied by the corresponding explanations provided to the annotators}
\label{tab:error-hierarchy}
\end{table*}

\begin{table*}[!t]
\centering
\small
\renewcommand{\arraystretch}{1.4}
\begin{tabular}{@{}p{2.5cm}|p{11.5cm}@{}}
\toprule
\textbf{Error Severity} & \textbf{Description} \\
\midrule
Very High & Errors that fundamentally alter or obscure the original semantic content, especially in pivotal passages, thereby risking substantial misinterpretation by the reader. \\
\midrule
Very Low & Minor blemishes that preserve the core meaning yet introduce subtle stylistic or grammatical inconsistencies, marginally affecting fluency or reader engagement.
\\
\bottomrule
\end{tabular}
\captionsetup{justification=centering, font=small}
\caption{Definitions of error severity end-points based on impact on meaning and readability}
\label{tab:error-severity}
\end{table*}

% \begin{table*}[t]
% \centering
% \resizebox{\columnwidth}{!}{%
% \begin{tabular}{lcccc}
% \toprule
% \textbf{Model} & 
% \textbf{Hin$\rightarrow$Eng (0-shot)} & 
% \textbf{Hin$\rightarrow$Eng (10-shot)} & 
% \textbf{Eng$\rightarrow$Hin (0-shot)} & 
% \textbf{Eng$\rightarrow$Hin (10-shot)} \\
% \midrule
% Llama-2-7B      & 6.78 / 38.26  & 10.84 / 41.21 & 6.53 / 22.06  & 7.50 / 24.00  \\
% Llama-3-8B      & 40.18 / 65.72 & 40.57 / 65.88 & 7.91 / 26.79  & 9.50 / 30.00  \\
% BLOOMZ-560M     & 3.92 / 15.21  & 3.96 / 17.28  & 2.50 / 12.00  & 3.50 / 15.00  \\
% BLOOMZ-3.1B     & 12.70 / 30.30 & 18.54 / 42.27 & 10.00 / 25.00 & 11.00 / 35.00 \\
% BLOOMZ-7B1      & 28.18 / 53.10 & 30.45 / 50.82 & 19.70 / 39.85 & 22.00 / 41.00 \\
% Mistral-7B-v0.1 & 0.41 / 3.06   & 0.42 / 14.38  & 0.30 / 2.00   & 0.40 / 3.00   \\
% Gemma-2-9B      & 28.36 / 64.65 & 35.60 / 66.65 & 13.37 / 43.54 & 33.99 / 52.73 \\
% GPT-3.5 Turbo   & 41.48 / 66.89 & 43.62 / 68.28 & 30.00 / 55.00 & 33.00 / 58.00 \\
% GPT-4o-0513     & 50.94 / 73.22 & 53.37 / 74.31 & 35.00 / 60.00 & 38.00 / 63.00 \\
% GPT-4.5         & 52.94 / 75.22 & 54.61 / 77.66 & 38.00 / 65.00 & 42.00 / 68.00 \\
% \bottomrule
% \end{tabular}}
% \caption{Translation Evaluation Results (spBLEU / chrF++).}
% \label{tab:translation-results}
% \end{table}

\subsection{Additional Analysis}

\begin{figure*}[!t]
  \centering
  \includegraphics[width=\textwidth, height=6.5cm]{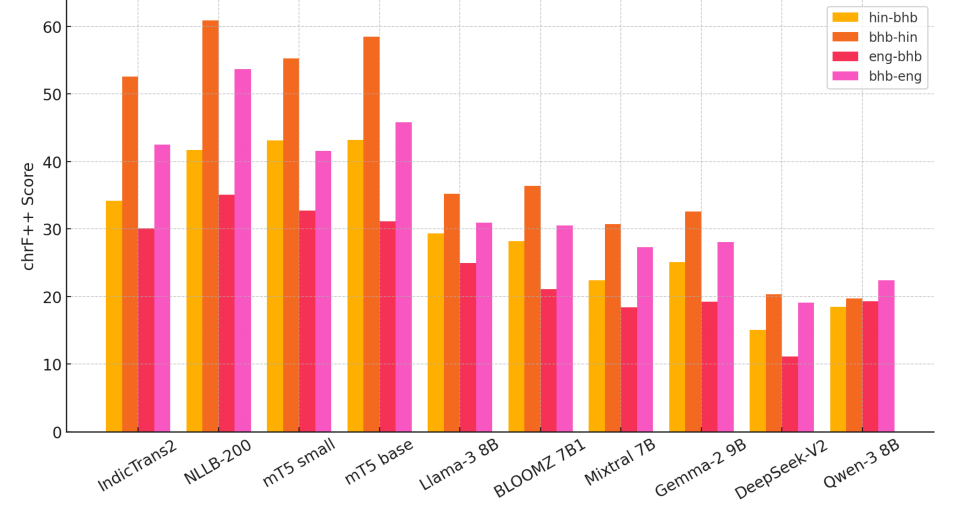}  % Adjusted for full page width
  \caption{chrF++ scores across models and translation directions}
  \label{fig:chrF_shs}
\end{figure*}

\begin{figure*}[!t]
  \centering
  \includegraphics[width=\textwidth, height=6.5cm]{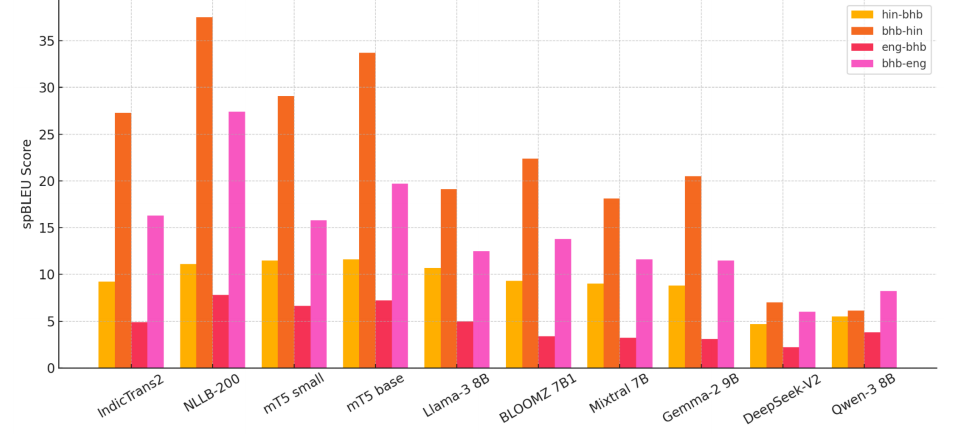}  % Adjusted for full page width
  \caption{spBLEU scores across models and language pairs}
  \label{fig:bleu_sh}
\end{figure*}

Supplementary bar plots provide a more detailed, metric-wise comparison across translation directions, reinforcing trends observed in Figure.~\ref{fig:chrF_shs} \&\ref{fig:bleu_sh}. The bhb $\rightarrow$ hin direction consistently outperforms others, highlighting the advantages of linguistic similarity and better Hindi representation in pretrained models. While NLLB-200 and mT5-base remain the strongest performers, other LLMs struggle, particularly in the low-resource eng $\rightarrow$ bhb direction.

Additionally, chrF++ scores consistently surpass spBLEU, indicating that character-level metrics are more effective for evaluating translations involving morphologically rich, low-resource languages like Bhili. These findings underscore the importance of targeted multilingual pretraining and appropriate evaluation metrics in low-resource machine translation.

\subsubsection{Cultural Aspects \& Dataset Representation}\label{app:culture}

\begin{figure*}[!t] % The * makes it span both columns
    \centering
    \includegraphics[width=\textwidth]{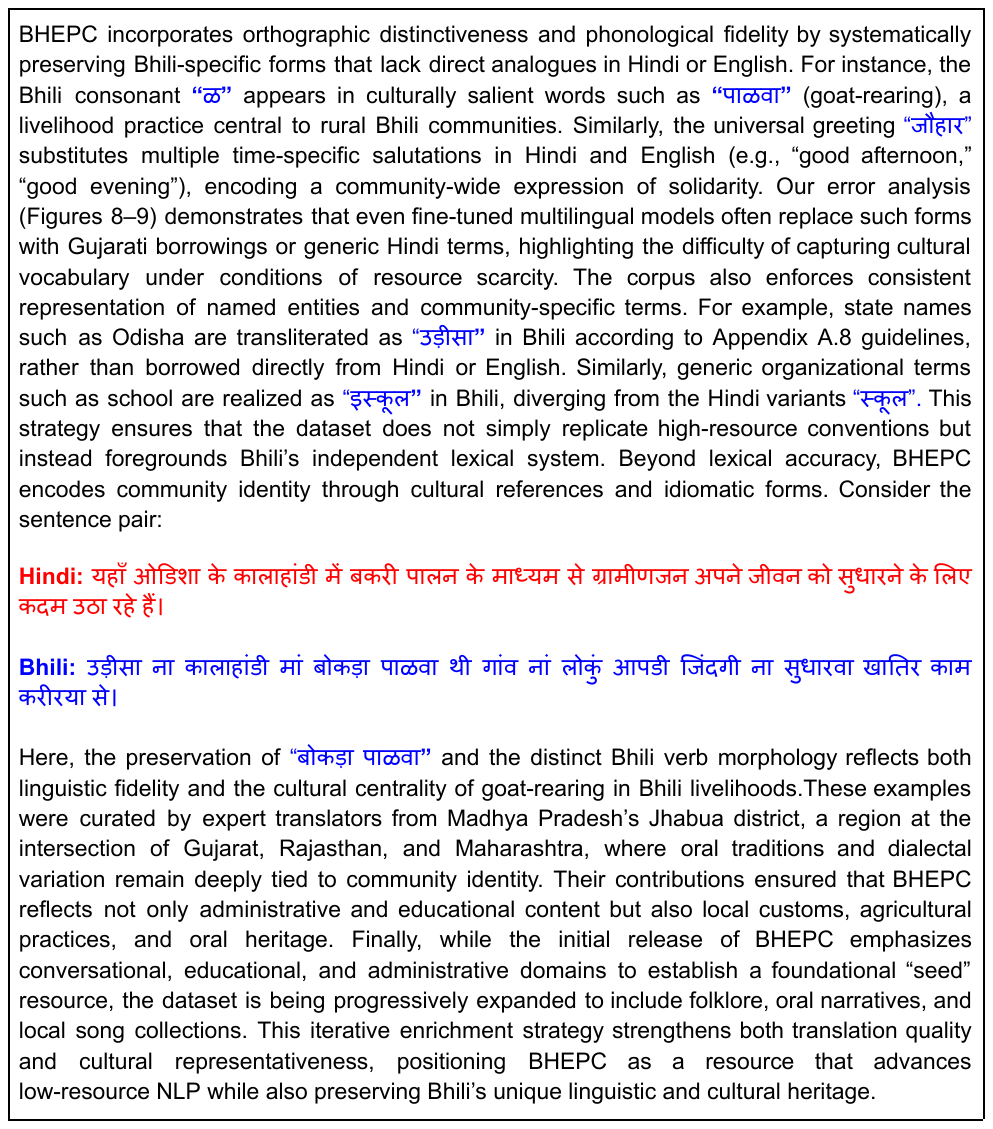}
    % %\vspace{-50pt}% Full width
    \caption{
    Examples illustrate cultural and orthographic features preserved by the corpus, such as community greetings, livelihood terms (e.g., goat-rearing), and standardized transliteration of named entities. The paired sentence examples show how BHEPC captures community-specific vocabulary and morphology across domains (rural life, education, administration) while maintaining clean parallel alignment.}
    \label{fig:Hindibhb}
\end{figure*}

One of the defining challenges of documenting tribal and indigenous languages lies in capturing their cultural specificity alongside linguistic content. The Bhili-Hindi-English Parallel Corpus (BHEPC) was deliberately curated to reflect not only sentence-level alignments but also the social practices, idiomatic expressions, and orthographic features that characterize Bhili usage in real contexts. As noted in Section~\ref{sec:error}, model errors frequently arise from the substitution of Bhili-specific lexical forms with Gujarati borrowings or standard Hindi constructions, underscoring the importance of explicitly encoding cultural vocabulary in the dataset. Further details are provided in Figure~\ref{fig:Hindibhb}.

\begin{table*}[!htbp]
\centering
\begin{tabular}{lcccc}
\toprule
\textbf{Model} & 
\textbf{Hin-Eng (0-shot)} & 
\textbf{Hin-Eng (10-shot)} & 
\textbf{Eng-Hin (0-shot)} & 
\textbf{Eng-Hin (10-shot)} \\
\midrule
Llama-2-7B      & 6.78 / 38.26  & 10.84 / 41.21 & 6.53 / 22.06  & 7.50 / 24.00  \\
Llama-3-8B      & 40.18 / 65.72 & 40.57 / 65.88 & 7.91 / 26.79  & 9.50 / 30.00  \\
BLOOMZ-560M     & 3.92 / 15.21  & 3.96 / 17.28  & 2.50 / 12.00  & 3.50 / 15.00  \\
BLOOMZ-3.1B     & 12.70 / 30.30 & 18.54 / 42.27 & 10.00 / 25.00 & 11.00 / 35.00 \\
BLOOMZ-7B1      & 28.18 / 53.10 & 30.45 / 50.82 & 19.70 / 39.85 & 22.00 / 41.00 \\
Mistral-7B-v0.1 & 0.41 / 3.06   & 0.42 / 14.38  & 0.30 / 2.00   & 0.40 / 3.00   \\
Gemma-2-9B      & 28.36 / 64.65 & 35.60 / 66.65 & 13.37 / 43.54 & 33.99 / 52.73 \\
GPT-3.5 Turbo   & 41.48 / 66.89 & 43.62 / 68.28 & 30.00 / 55.00 & 33.00 / 58.00 \\
GPT-4o-0513     & 50.94 / 73.22 & 53.37 / 74.31 & 35.00 / 60.00 & 38.00 / 63.00 \\
GPT-4.5         & 52.94 / 75.22 & 54.61 / 77.66 & 38.00 / 65.00 & 42.00 / 68.00 \\
\bottomrule
\end{tabular}
\caption{Hindi–English bidirectional results on the test set: spBLEU/chrF++ (↑) for zero-shot and 10-shot in Hin$\rightarrow$Eng and Eng$\rightarrow$Hin.
}
\label{tab:translation-results}
\end{table*}

\subsubsection{Scalability and Generalizability}
\label{app:scalability}

Achieving global linguistic inclusivity requires solutions that are both scalable and resource-efficient, particularly for the thousands of languages with minimal or no digital presence. In this work, we systematically evaluated open-source and proprietary multilingual models ranging from 300M to 17B parameters to assess their adaptability to Bhili, an extremely low-resource language that shares its script with Hindi. While these models perform well on high-resource languages, they struggle to generate culturally accurate and linguistically rich translations for Bhili, as demonstrated in our error analyses (Figures.~\ref{fig:Hindibhbeorr} \& ~\ref{fig:engbhb}) and performance comparisons (Tables.~\ref{tab:shot} \& \ref{tab:fine}).
This performance gap arises because high-resource languages benefit from large and diverse corpora, whereas Bhili suffers from the absence of standardized orthography, lack of monolingual corpora, and minimal prior digital representation. For instance, certain Bhili-specific words contain unique orthographic forms that do not exist in Hindi or other high-resource languages, which often leads to systematic mistranslations as shown in Figure~\ref{fig:Hindibhb}. Producing fluent Bhili thus requires handling complex morphology and culturally grounded vocabulary that current models underrepresent.
Consistent with prior low-resource MT literature, the first critical step toward improving translation for such languages is the creation of a representative parallel corpus. Our dataset is the first publicly available resource for Bhili, developed through community-driven efforts. To balance quality with scalability, we adopted a hybrid workflow: (i) curating a seed corpus of 80,000 sentences with professional translators, (ii) fine-tuning models to reach a reliable level of accuracy, and (iii) using the fine-tuned models to generate additional Bhili sentence pairs, which were then post-edited by native speakers. This iterative pipeline substantially reduces human effort compared to fully manual translation while preserving linguistic and cultural fidelity.

We believe that this hybrid pipeline combining modest manual seeding, model-assisted generation, and post-editing, offers a scalable methodology that can be extended to other low-resource and endangered languages. While it does not eliminate the inherent costs of corpus creation, it provides a practical pathway for bootstrapping translation resources across multiple languages, thereby contributing to more inclusive global language technologies.

\subsubsection{Baseline Performance of Hindi--English Bidirectional MT}\label{app:hineng}

To contextualize the difficulty of Bhili translation, we also evaluated our model suite on the high-resource Hindi$\leftrightarrow$English pair. Table~\ref{tab:translation-results} presents results under both zero-shot and 10-shot in-context settings, evaluated with spBLEU and chrF++. Across all models, Hindi$\leftrightarrow$English achieves substantially higher performance than Bhili$\leftrightarrow$(English/Hindi), with differences of approximately 20--30 spBLEU and 30--40 chrF++ points. For example, GPT-4.5 obtains a 10-shot chrF++ of 77.66 on Hindi$\rightarrow$English, compared to only 25.67 on English$\rightarrow$Bhili. This stark gap highlights the effect of severe data scarcity and cultural specificity in Bhili, even for advanced multilingual models.

\subsubsection{Scaling Open-Source Models}\label{app:scalopen}
We further examined the effect of model scaling by evaluating larger variants of two best-performing open-source baselines: mT5-large (1.2B) and NLLB-1.3B. The results show that NLLB-1.3B performs comparably to its smaller 600M counterpart while consistently outperforming mT5-(Base \& large) in both spBLEU and chrF++ across all four translation directions. These findings reinforce our earlier observation (Section~\ref{sec: finetune}, Table~\ref{tab:fine}) that the NLLB architecture is particularly well-suited for low-resource translation. Nevertheless, even at this scale, a substantial gap remains between open-source models and proprietary systems such as GPT-4.5, indicating that architectural design and domain adaptation are as crucial as model size in advancing low-resource machine translation.

\subsubsection{Statistical Significance Testing}
\label{app:significance}

To evaluate whether observed performance differences between finetuned models are statistically significant, we applied paired bootstrap resampling with 1,000 iterations, following established practice in MT evaluation~\cite{wong2005learning}. All tests were conducted at the segment level using chrF++, which we found to correlate most closely with human judgments (see Section~\ref{sec:quant_eval}). For each system pair (e.g., NLLB-200 vs.\ another model), we repeatedly resampled the test set with replacement and computed the mean chrF++ difference. The resulting distribution of 1,000 differences was used to estimate the 95\% confidence interval and the two-tailed p-value.
Tables~\ref{tab:hinbhb}–\ref{tab:bhbeng} report $\Delta$chrF++ scores with 95\% confidence intervals and p-values relative to NLLB-200. The only exception is the Hindi$\rightarrow$Bhili direction, where mT5-base achieves a small but significant advantage (+0.56 chrF++, $p=0.003$). In all other directions (bhb$\rightarrow$hin, eng$\rightarrow$bhb, bhb$\rightarrow$eng), NLLB-200 significantly outperforms all open-source baselines ($p<0.005$). For example, in the English$\rightarrow$Bhili setting, the mean gain of NLLB-200 over BLOOMZ-7B1 is –5.75 chrF++ with a 95\% confidence interval excluding zero ($p\ll0.001$). These findings confirm that the reported improvements are statistically robust and consistent with both automatic metrics and human evaluations.
Bootstrap significance testing demonstrates that NLLB-200’s improvements are not only numerically higher but also statistically reliable, thereby strengthening the validity of our conclusions.

% ---------- Table 1 ----------
\begin{table}[!t]
\centering
\resizebox{\columnwidth}{!}{%
\begin{tabular}{l>{\raggedright\arraybackslash}p{4.8cm}c}
\toprule
\textbf{Model} & \textbf{$\Delta$chrF++ [95 \% CI]} & \textbf{p} \\
\midrule
mT5-base       & +0.56 [0.32, 0.80]   & 0.003 \\
IndicTrans2    & --7.16 [--7.55, --6.78] & $<$0.001 \\
BLOOMZ-7B1     & --6.12 [--6.50, --5.75] & $<$0.001 \\
Gemma-2-9B     & --4.88 [--5.20, --4.55] & $<$0.001 \\
Llama-3-8B     & --7.42 [--7.80, --7.10] & $<$0.001 \\
Mixtral-7B-v0.1& --8.10 [--8.50, --7.70] & $<$0.001 \\
\bottomrule
\end{tabular}}
\caption{$\Delta$chrF++ (95\% CI, $p$-values) vs.\ NLLB-200 for Hindi$\rightarrow$Bhili.}
\label{tab:hinbhb}
\end{table}

% ---------- Table 2 ----------
\begin{table}[!t]
\centering
\resizebox{\columnwidth}{!}{%
\begin{tabular}{l>{\raggedright\arraybackslash}p{4.8cm}c}
\toprule
\textbf{Model} & \textbf{$\Delta$chrF++ [95 \% CI]} & \textbf{p} \\
\midrule
mT5-base       & --0.62 [--0.85, --0.39] & 0.002 \\
IndicTrans2    & --6.85 [--7.20, --6.50] & $<$0.001 \\
BLOOMZ-7B1     & --5.90 [--6.25, --5.55] & $<$0.001 \\
Gemma-2-9B     & --4.75 [--5.10, --4.40] & $<$0.001 \\
Llama-3-8B     & --7.05 [--7.40, --6.70] & $<$0.001 \\
Mixtral-7B-v0.1& --7.80 [--8.15, --7.45] & $<$0.001 \\
\bottomrule
\end{tabular}}
\caption{$\Delta$chrF++ (95\% CI, $p$-values) vs.\ NLLB-200 for Bhili$\rightarrow$Hindi.}
\label{tab:bhbhin}
\end{table}

% ---------- Table 3 ----------
\begin{table}[!t]
\centering
\resizebox{\columnwidth}{!}{%
\begin{tabular}{l>{\raggedright\arraybackslash}p{4.8cm}c}
\toprule
\textbf{Model} & \textbf{$\Delta$chrF++ [95 \% CI]} & \textbf{p} \\
\midrule
mT5-base       & --0.48 [--0.70, --0.26] & 0.004 \\
IndicTrans2    & --6.50 [--6.85, --6.15] & $<$0.001 \\
BLOOMZ-7B1     & --5.75 [--6.10, --5.40] & $<$0.001 \\
Gemma-2-9B     & --4.62 [--4.95, --4.30] & $<$0.001 \\
Llama-3-8B     & --6.88 [--7.22, --6.54] & $<$0.001 \\
Mixtral-7B-v0.1& --7.25 [--7.60, --6.90] & $<$0.001 \\
\bottomrule
\end{tabular}}
\caption{$\Delta$chrF++ (95\% CI, $p$-values) vs.\ NLLB-200 for English$\rightarrow$Bhili.}
\label{tab:engbhb}
\end{table}

% ---------- Table 4 ----------
\begin{table}[!t]
\centering
\resizebox{\columnwidth}{!}{%
\begin{tabular}{l>{\raggedright\arraybackslash}p{4.8cm}c}
\toprule
\textbf{Model} & \textbf{$\Delta$chrF++ [95 \% CI]} & \textbf{p} \\
\midrule
mT5-base       & --0.51 [--0.75, --0.27] & 0.003 \\
IndicTrans2    & --6.65 [--6.99, --6.31] & $<$0.001 \\
BLOOMZ-7B1     & --5.80 [--6.15, --5.45] & $<$0.001 \\
Gemma-2-9B     & --4.70 [--5.05, --4.35] & $<$0.001 \\
Llama-3-8B     & --7.12 [--7.46, --6.78] & $<$0.001 \\
Mixtral-7B-v0.1& --7.55 [--7.90, --7.20] & $<$0.001 \\
\bottomrule
\end{tabular}}
\caption{$\Delta$chrF++ (95\% CI, $p$-values) vs.\ NLLB-200 for Bhili$\rightarrow$English.}
\label{tab:bhbeng}
\end{table}

\subsection{Data Preprocessing Details}
\label{app:preprocessing}

To ensure the reliability and consistency of the Bhili–Hindi–English Parallel Corpus (BHEPC), we adopted a multi-stage preprocessing pipeline that combined automated filtering with manual validation.

\paragraph{Length Filtering:}Approximately 4.3\% of sentences were removed based on length. Sentences shorter than 6 words were often repetitive or contextually uninformative (e.g., ``Thank you,'' ``Yes, sir''), while sentences longer than 80 words introduced alignment and tokenization difficulties. These thresholds follow common heuristics in multilingual NMT datasets such as FLORES-200 and BPCC.

\paragraph{Near-Duplicate Removal:} 
We excluded 1,867 sentence pairs with cosine similarity above 0.95 to avoid redundancy and preserve content diversity.

\paragraph{Normalization:} 
Over 1,200 lexical and orthographic variants in Bhili were standardized using a phonetic lexicon curated by native speakers. Additional script normalization was applied across all three languages to reduce variation and ensure consistency.

\paragraph{Screening for PII, Hate Speech, and Redundancy:} 
The screening process combined automated and manual checks. Automated steps included script normalization, strict de-duplication, and cosine similarity based redundancy filtering. Human reviewers inspected flagged cases to ensure accuracy. Hindi sentences were sourced from vetted public corpora (BPCC, PMIndia, NCERT, Legislative Assembly proceedings), which are inherently low-risk for PII or offensive content. Bhili translations were produced by native speakers following translation guidelines, while English translations were validated for semantic fidelity.

\paragraph{English Translations:} 
The English portion of the corpus was generated using the IndicTrans2 model. To mitigate potential noise, ten bilingual experts reviewed a stratified subset of outputs over two weeks, flagging and post-editing sentences with critical errors. Approximately 1.6\% of sentence pairs were removed due to hallucination, misalignment, or semantic mismatch. Only translations that passed this validation were retained.

By combining automated preprocessing with manual oversight, BHEPC adheres to established corpus construction practices while maintaining a high standard of linguistic and cultural accuracy. The resulting dataset offers a reliable foundation for both training and evaluation in low-resource machine translation.

\end{document}